\documentclass{article}

\usepackage{arxiv}

\usepackage[utf8]{inputenc} 
\usepackage[T1]{fontenc}    
\usepackage{hyperref}       
\usepackage{url}            
\usepackage{multirow}
\usepackage{siunitx}
\usepackage[table,xcdraw]{xcolor}
\usepackage{booktabs}       
\usepackage{amsfonts}       
\usepackage{nicefrac}       
\usepackage{rotating}
\usepackage{float}
\usepackage{microtype}      %
\usepackage{threeparttable}
\usepackage{amsmath} 
\usepackage{lipsum}
\usepackage{graphicx}
\usepackage{subcaption}
\usepackage{caption}

\graphicspath{ {./images/} }

\title{MoDyGAN: Combining Molecular Dynamics With GANs to Investigate Protein Conformational Space}

\author{
  Jingbo Liang \\
  Department of Computer Science\\
  University of New Mexico\\
  Albuquerque, NM 87106 \\
  \texttt{liangjingbo@unm.edu}
  \And
  Bruna Jacobson\thanks{Corresponding author: \texttt{bjacobson@unm.edu}} \\
  Department of Computer Science\\
  University of New Mexico\\
  Albuquerque, NM 87106 \\
  \texttt{bjacobson@unm.edu}
}

\begin{document}
\maketitle
\begin{abstract}
Extensively exploring protein conformational landscapes remains a major challenge in computational biology due to the high computational cost involved in dynamic physics-based simulations. In this work, we propose a novel pipeline, MoDyGAN, that leverages molecular dynamics (MD) simulations and generative adversarial networks (GANs) to explore protein conformational spaces. MoDyGAN contains a generator that maps Gaussian distributions into MD-derived protein trajectories, and a refinement module that combines ensemble learning with a dual-discriminator to further improve the plausibility of generated conformations. Central to our approach is an innovative representation technique that reversibly transforms 3D protein structures into 2D matrices, enabling the use of advanced image-based GAN architectures. We use three rigid proteins to demonstrate that MoDyGAN can generate plausible new conformations. We also use deca-alanine as a case study to show that interpolations within the latent space closely align with trajectories obtained from steered molecular dynamics (SMD) simulations. Our results suggest that representing proteins as image-like data unlocks new possibilities for applying advanced deep learning techniques to biomolecular simulation, leading to an efficient sampling of conformational states. Additionally, the proposed framework holds strong potential for extension to other complex 3D structures.
\end{abstract}


\section{Introduction}
Proteins are dynamic molecules that fold into specific 3D structures that are crucial for their function. Prediction of physical and chemical properties of proteins is traditionally done via Molecular Dynamics (MD) simulations that provide valuable insight into protein dynamic behavior at the atomic scale and at very fine temporal resolution, complementing experimental approaches. However, MD simulations face the issue of extensive sampling, as time and computational resources limit their ability to explore the whole conformational space of proteins ~\cite{yang2019enhanced}. Due to these limitations of traditional enhanced sampling methods~\cite{yang2019enhanced}, such as collective variables (CVs)-based \cite{valsson2016enhancing,valsson2014variational} and CVs-free \cite{sugita1999replica,yang2015thermodynamics} sampling methods, it is imperative to develop more \emph{intelligent} methods to overcome this problem.

Recently, unsupervised deep learning-based sampling methods, such as variational autoencoders (VAEs)~\cite{kingma2013auto}, have shown significant promise in enhancing extensive sampling~\cite{degiacomi2019coupling,tian2021explore}. However, the potential of generative adversarial networks (GANs)~\cite{goodfellow2014generative}, as another advanced unsupervised deep learning technique, remains unexplored. Compared to VAEs, GANs have inherent advantages, such as producing more detailed outputs~\cite{rosca2017variational} for exploration of small changes in protein dynamics, and more interpretable latent spaces \cite{schon2022interpreting} that provide a reduced-dimensional representation that captures essential patterns of protein conformation and dynamics.

In this paper, we propose a novel GAN-based extensive sampling approach, which combines GANs with MD simulations to accelerate the exploration of the protein conformational landscape, allowing deeper insight into a protein's dynamic processes. \textbf{We hypothesize that GAN models, which are trained using conformation data derived from MD simulations, can produce new, plausible conformations that complement existing conformations.} Our study represents an opportunity to fill a gap in applying GANs in protein dynamics studies, with important implications for drug design and bioengineering. Our work makes the following contributions:
\begin{itemize}

\item We introduce an innovative, orientation-independent image-based representation method for protein conformations. This approach facilitates straightforward reversible translations between 2D matrices and 3D structures and can be effectively applied in established image-based GAN techniques.

\item We propose a novel pipeline that is capable of generating new and plausible protein conformations with variable and extended backbone lengths. Central to this pipeline is a refinement module that employs ensemble learning and a dual discrimination framework.

\item This paper also addresses the gap in applying GANs to protein dynamics research, offering significant potential benefits for broader scientific and bioengineering applications.

\end{itemize}

\section{Related Work}
Unsupervised deep learning-based sampling methods have shown promise in simulating macromolecular dynamics, although at an early stage \cite{hoseini2021generative}. For example, \cite{degiacomi2019coupling} developed an autoencoder model to explore the conformational space of proteins. It employs an encoder that maps MD simulation-derived conformations into a latent space and a decoder that reconstructs the protein conformation back from this latent space. 
While their approach demonstrated encouraging results in capturing molecular conformational variability, compared to GANs, autoencoder-based models have several inherent limitations, such as generating less detailed output \cite{rosca2017variational} and less interpretable latent spaces \cite{schon2022interpreting}.

GANs are powerful machine learning frameworks for generative artificial intelligence \cite{goodfellow2014generative}. While they have been widely adopted in image generation tasks, their application to molecular simulations remains limited \cite{hoseini2021generative}. 
To date, the only notable application of GANs in this field is Targeted Adversarial Learning Optimized Sampling (TALOS) \cite{zhang2019targeted}. 
Although it has shown promise in simple systems, such as the SN2 reaction and alanine dipeptide, broader development of GAN-based sampling methods remains a significant opportunity.

The primary barrier to applying GANs to protein structure generation and dynamic modeling lies in designing scalable and generalizable protein representations. Currently, there are four chemical representations. The first is graph-based representations, combining adjacency matrices with node features~\cite{macedo2024medgan,de2018molgan}. They fit naturally with graph-based GANs but fail to fully capture 3D conformations and scale poorly with protein size.  RamaNet \cite{sabban2019ramanet} adopts torsion angle-based sequential representations to generate helical structures. However, as this format is not naturally compatible with standard GAN architectures, specialized frameworks need to be developed, such as the LSTM-based GAN~\cite{sabban2019ramanet}, posing challenges for broader generalization across GAN models. The third is using explicit 3D atomic coordinates \cite{degiacomi2019coupling}, which are inherently orientation-dependent \cite{li2017protein},
 thus complicating learning. While 3D atomic coordinates are widely used in VAEs, they remain challenging to generalize in GANs. The last representation is pairwise distance matrices, which have been adopted in several studies \cite{anand2018generative, anand2019fully}. These grayscale, image-like matrices are orientation-invariant and enable the application of established image-based GAN architectures, such as Deep Convolutional Generative Adversarial Networks (DCGANs)~\cite{radford2015unsupervised}, for the generation of structural fragments. However, as distance information alone often leads to ambiguities, reconstructing 3D coordinates from distance matrices requires additional reconstruction methods~\cite{li2017protein,anand2018generative,anand2019fully}. Besides, GANs trained only on fixed-length fragments struggle to generalize to proteins with varying lengths or more complex topologies\cite{anand2018generative, anand2019fully,li2017protein}, highlighting a scalability limitation. The image-like representation offers a compelling solution due to its compatibility with image-based GAN architectures and support for efficient generative modeling, although further improvements are needed to enhance its scalability and the preservation of essential spatial information.

Given an improved image-like representation suitable for image-based GANs, the next obstacle lies in applying advanced GAN variants capable of capturing finer structural details. Previous studies have predominantly utilized DCGANs~\cite{radford2015unsupervised} for protein structure generation~\cite{anand2018generative,anand2019fully,li2017protein}. However, DCGANs are prone to mode collapse, limiting their ability to learn the true distribution of protein conformations. Furthermore, their inability to generate high-resolution outputs restricts outputs to be oligopeptides or fragments of proteins. For example, \cite{anand2018generative} used DCGANs to generate protein backbone structures but could only produce fragments up to 128 residues. Fortunately, more advanced variants of GANs have been developed to address these problems. Wasserstein GAN with Gradient Penalty (WGAN-GP) \cite{gulrajani2017improved} solves mode collapse by incorporating Lipschitz continuity constraints during training. ProGAN~\cite{karras2017progressive} outputs high-resolution image by progressively growing both the generator and discriminator networks and incorporating several innovative techniques, such as fade-in transitions, equalized learning rates, pixelwise feature normalization, and adding a minibatch standard deviation layer in the discriminator. 

Lastly, GAN outputs often contain structural errors that require further refinement. Although some studies apply no explicit refinement~\cite{anand2019fully}, most incorporate a refinement module, such as an additional trained neural network~\cite{anand2018generative}. By treating protein conformation as an \emph{image}, erroneous conformations can be viewed as a \emph{noisy image}. Thus, we view the refinement process as an image-denoising task. This fits well with Pix2Pix model~\cite{isola2017image}, a conditional GAN designed to generate outputs based on input data. However, Pix2Pix applies uniform attention across all pixels, leading to over- or under-refinement in different regions. Specifically, rigid secondary structures, such as $\alpha$-helices and $\beta$-sheets, typically exhibit fixed patterns, whereas flexible regions like loops and turns do not. Optimizing Pix2Pix for rigid structures leads to over-refinement of flexible regions, reducing their natural flexibility, while optimizing for flexible regions results in under-refinement of rigid structures, weakening their stability. Thus, further improvements are necessary to address this imbalance.

\section{Methods}
\begin{figure}
    \centering    
    \includegraphics[width=1.0\linewidth]{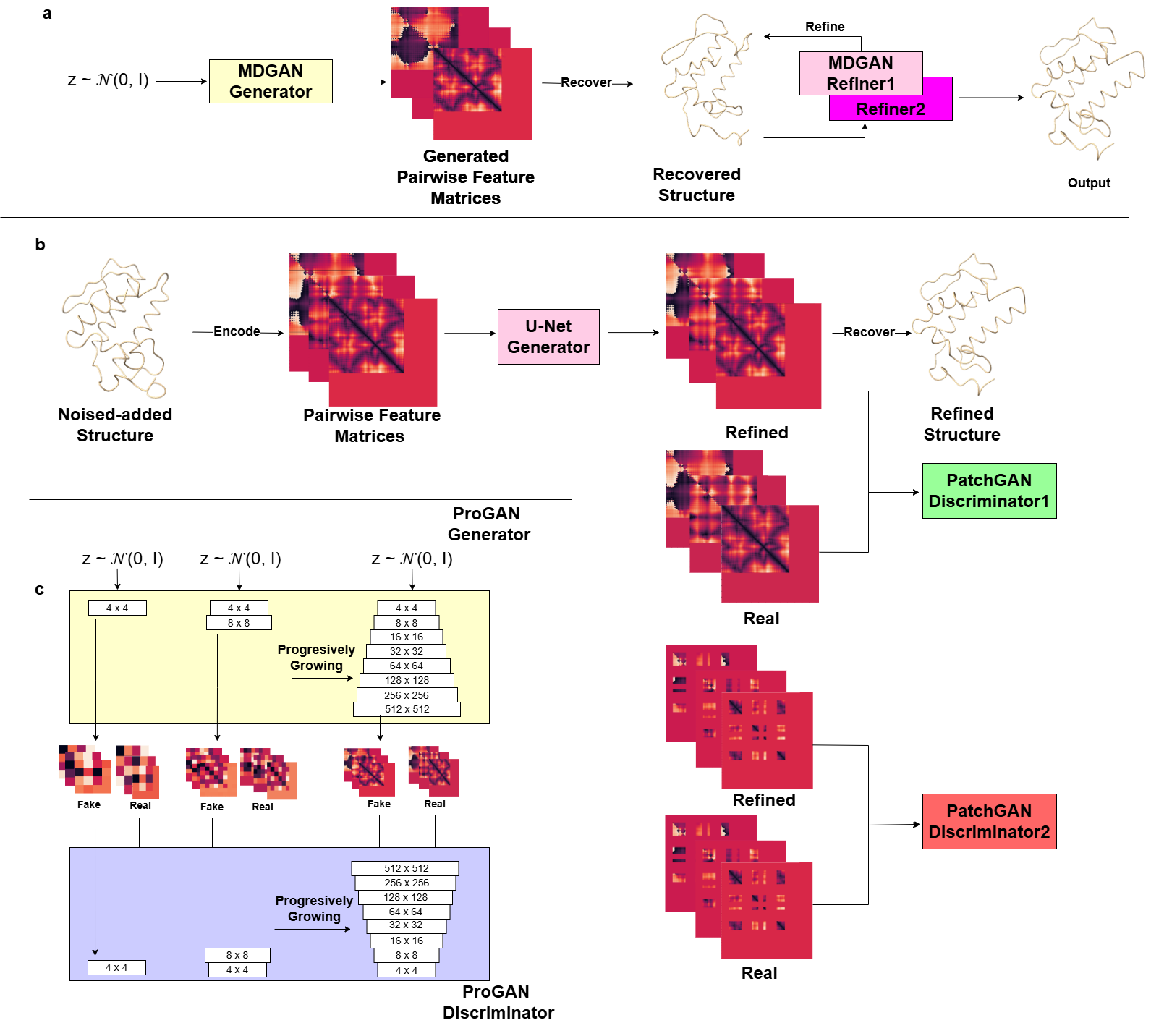}
    \caption{~\textbf{(a) Pipeline.} The MoDyGAN generator generates pairwise feature matrices, which are then recovered in \( O(n^2) \) time. 
    The refinement module combines ensemble learning with a dual-discriminator to produce the final outputs.~\textbf{(b) Refiner.} Modified Pix2Pix~\cite{isola2017image} architecture. The generator takes noise-added pairwise feature matrices derived from noisy conformations and maps them to their clean counterparts. Discriminator1 is trained to distinguish between the generator’s denoised outputs and the corresponding ground-truth matrices, thereby enforcing overall structural consistency. Discriminator2 focuses on secondary structure regions by using masked feature matrices from both ground-truth and generated conformations. It is trained to distinguish between them, guiding the generator to better preserve secondary structural accuracy. The two best-performing generators are subsequently incorporated in the dual refinement module within our pipeline.~\textbf{(c) Generator.} A standard ProGAN~\cite{karras2017progressive} model is trained to map Gaussian latent samples to real protein conformations. Once trained, the ProGAN generator serves as the MoDyGAN generator in our pipeline to generate initial pairwise feature matrices. 
    }

    \label{fig:pipelineMD}
\end{figure}

As depicted in Figure \ref{fig:pipelineMD}, our approach comprises three sequential stages: first, we \textbf{generate} pairwise feature matrices that encode the protein backbone; next, we \textbf{recover} the corresponding 3D coordinates; and finally, we \textbf{refine} these coordinates to correct local backbone inaccuracies.

\subsection{Feature Maps Generation}
We designed novel 2D pairwise feature matrices to represent protein structures~(see Supplementary Figure~S~\ref{fig:proteinRepresentation}). This representation treats each protein conformation as an \emph{image} composed of three distinct channels for each coordinate of the pairwise distance vector between backbone atoms in spherical coordinates ($d$, $\theta$, and $\phi$). For a structure with $n$ backbone atoms, we extract the 3D coordinates and process them into an image with size $n \times n \times 3$ as illustrated in Supplementary Figure~S~\ref{fig:proteinRepresentation}. These matrices are normalized and resized by zero-padding to standard dimensions (e.g., 64, 128, 256, or 512) to ensure compatibility with downstream GAN models. This representation is orientation independent and implicitly encodes structural relationships 
while enabling efficient reconstruction of protein backbones.

\subsection{Generation}

As illustrated in Figure \ref{fig:pipelineMD}~a, MoDyGAN's generator receives an input vector sampled from the Gaussian distribution and outputs pairwise feature matrices that represent protein backbone structures. To achieve this, we train a ProGAN~\cite{karras2017progressive,persson2020progan} model (see Figure \ref{fig:pipelineMD}~c), whose generator is optimized to create realistic matrices, and discriminator is trained to differentiate authentic matrices from generated ones. The trained generator from ProGAN serves as the generator in our pipeline.

\subsection{Recovery}
Because the pairwise feature representation preserves all information needed for 3D reconstruction of the protein backbone, the recovery process is straightforward. We first denormalize and remove padding from the generated matrices. The $i$-th row in the resulting matrix describes the conformation in spherical coordinates centered at atom $i$. Let $\mathbf{S}_i \in \mathbb{R}^{n \times 3}$ denote the spherical coordinate matrix derived from the $i$-th row. Each entity of $\mathbf{S}_i$, representing an atom $\mathbf{a}_j$ with spherical coordinates $(r_j, \phi_j, \theta_j)$, is transformed into Cartesian coordinates using the standard mapping: $x_j = r_j \sin(\phi_j) \cos(\theta_j)$, $y_j = r_j \sin(\phi_j) \sin(\theta_j)$, and $z_j = r_j \cos(\phi_j)$ (see Supplementary Text~D for more details).

By repeating this process for all $i = 1, \dots, n$, we obtain $n$ conformations, each centered on a different backbone atom. These 3D coordinate sets are stacked and averaged to produce a final recovered structure. The whole recovery process takes \( O(n^2) \) time.  

\subsection{Refinement}
After recovering conformations, we observe local backbone inaccuracies, particularly around secondary structure regions. To address these errors, we trained a modified Pix2Pix with dual discriminators and employ an ensemble learning strategy, in which two refiners are applied iteratively to refine the outputs (Figure \ref{fig:pipelineMD}a). The original Pix2Pix framework~\cite{isola2017image,persson2020Pix2Pix} consists of a generator ($G$) and a global discriminator ($D_1$) that evaluates overall matrix realism. Here we introduce a key modification: an additional discriminator ($D_2$) to specifically address structural inaccuracies within secondary structure regions (Figure~\ref{fig:pipelineMD}b). Let $x$ denote the noisy input pairwise feature matrix, $y$ be the corresponding ground-truth or clean matrix, and $G(x)$ the generator's output. We also define $\hat{x} $, $\hat{y}$, and $\hat{G(x)}$ as the secondary-structure–focused versions of $x$, $y$, and $G(x)$, respectively, where the non-secondary regions are masked out. We design the optimization objective as Equation~\ref{eq:objective}:

\begin{equation}
G^* = \arg \min_{G} \max_{D_1, D_2} \; \mathcal{L}(G, D_1, D_2)
\label{eq:objective}
\end{equation}

And the loss function is shown in Equation~\ref{eq:loss_total}: 

\begin{equation}
\begin{aligned}
\mathcal{L}(G, D_1, D_2) &= \mathcal{L}(G, D_1) + \mathcal{L}(G, D_2) + \lambda_{L1} \, \mathcal{L}_{L_1}(G) \\[6pt]
\mathcal{L}(G, D_1) &= \mathbb{E}_{x,y} \left[ \log D_1(x, y) \right] 
+ \mathbb{E}_x \left[ \log \left(1 - D_1(x, G(x)) \right) \right] 
+ \lambda_{\text{gp}} \, \mathrm{GP}(D_1) \\[6pt]
\mathcal{L}(G, D_2) &= \mathbb{E}_{\hat{x},\hat{y}} \left[ \log D_2(\hat{x}, \hat{y}) \right] 
+ \mathbb{E}_{\hat{x}} \left[ \log \left(1 - D_2(\hat{x}, \hat{G(x)}) \right) \right] 
+ \lambda_{\text{gp}} \, \mathrm{GP}(D_2) \\[6pt]
\mathcal{L}_{L_1}(G) &= \mathbb{E}_{x,y} \left[ \| y - G(x) \|_1 \right]
\end{aligned}
\label{eq:loss_total}
\end{equation}
\noindent where \( \lambda_{\text{gp}} = 10 \) and \( \lambda_{L1} = 1000 \).

$G$ is trained to minimize the loss function composed of adversarial feedback from two discriminators and an $L_1$ reconstruction loss. Each discriminator is trained to maximize its ability to distinguish real data from generated outputs using a standard adversarial loss, augmented with a gradient penalty term to promote stable training dynamics. $L_1$ loss is scaled by a large $\lambda_{L1}$, emphasizing the importance of accurate reconstruction structurally and pixel-wise close to the target. Operating in tandem, $D_1$ enforces global structural consistency, while $D_2$ focuses on preserving local secondary structure fidelity. In all rigid proteins, the two best-performing Pix2Pix generators are employed as dual refinement modules. For refinement of Ala\textsubscript{10}, we only used one discriminator as its highly flexible helix structure prevents applicability of another structure-aware discriminator. 

\section{Experiments}
To evaluate whether MoDyGAN can generate novel and physically plausible conformations within their conformational landscapes, we selected four proteins varying in size, structural complexity, and flexibility (see Supplementary Text A). Three of them are relatively rigid, characterized by stable secondary structures, including \textbf{phospholipase A\textsubscript{2}} (PDB ID: 1POA), \textbf{chain A of $\alpha$B-crystallin} (PDB ID: 2WJ7) and \textbf{alpha-like toxin LqH III} (PDB ID: 1BMR). Another system is \textbf{deca-alanine} (Ala\textsubscript{10}), a synthetic polypeptide composed of ten alanine residues. Ala\textsubscript{10} exhibits helix-coil transition behavior under steered molecular dynamics (SMD) (Supplementary Figure~S~\ref{supp:ala10}). Applying force to Ala\textsubscript{10} can eventually yield 20 distinct conformational states (see Supplementary Figure~S~\ref{fig:Ala_output}). MD simulations were performed to sample these system's conformational space, and the resulting data were then used to train MoDyGAN models to map a Gaussian latent distribution to the protein's conformational distribution. Finally, we evaluated our results through a model ablation study and comparisons with \cite{degiacomi2019coupling}. Data and code are available in https://github.com/UNMCoBSLab/MoDyGAN.

\subsection{Dataset}
MD simulations (see Supplementary Text~B) were conducted on these 5 systems:1POA, 2WJ7, 1BMR, and the initial and final states of Ala\textsubscript{10}. We generated over 30{,}000 conformations per system. All conformations were shuffled and then partitioned into four non-overlapping datasets:

\textit{Generator training dataset} contains 10{,}000 conformations to train the MoDyGAN generator. For the 3 rigid proteins, all conformations were randomly selected. For Ala\textsubscript{10}, 5{,}000 conformations were extracted from the MD simulation of the starting (folded) state and another 5{,}000 from the final (unfolded) state. 

\textit{Refiner training dataset} contains 10{,}000 pairs of noise-added and clean conformations used to train MoDyGAN refiners for denoising. This dataset was constructed by adding random displacements, uniformly sampled from $[-0.5, 0.5]\,\text{\AA}$, to the backbone atom coordinates of the generator training dataset. The noise-added conformations serve as inputs, with their corresponding clean conformations as targets.

\textit{Random forest (RF) dataset} is used to fit the RF classifiers and the \textit{validation set} assesses their performance (see Supplementary Text C for details).

\subsection{Implementation and Training}
Our program is implemented using PyTorch~\cite{paszke2019pytorch}, and all models are trained on an NVIDIA RTX A5000 GPU with 24~GB of memory. For both generator and refiner training, we used WGAN-GP loss function, whose coefficient of gradient penalty and drift penalty are 10 and 0.001, respectively. We employed Adam optimizer with $\beta_1 = 0.5$, $\beta_2 = 0.99$. Five discriminator updates were performed for each generator update. The MoDyGAN generator received a 100-dimensional vector sampled from a Gaussian distribution as input with learning rate of $1 \times 10^{-3}$, while refiners used paired samples from the \textit{refiner training dataset} with learning rate of $2 \times 10^{-4}$. To optimize memory efficiency, batch sizes and epoch are adjusted based on structure resolution  (see Supplementary Table~S~\ref{tab:conf}).

\subsection{Performance Evaluation}
All proteins were evaluated based on two criteria: \textbf{plausibility} and \textbf{novelty}. However, the evaluation strategy differs slightly between the three rigid proteins and Ala\textsubscript{10} (see Supplementary Text C for details).

For the three rigid proteins, \textbf{plausibility} was assessed by RF classification, bond and angle analysis, root mean square deviation (RMSD) and their backbone energy. For each system, the reference structure was defined as the initial conformation from its MD simulation trajectory. Generated conformations were then compared to this reference to calculate RMSD. Inspired by ~\cite{degiacomi2019coupling}, we also adopted a separately trained RF classifier to distinguish between native-like and non-native-like protein conformations. We used it to evaluate whether MoDyGAN-generated structures are plausible enough to deceive it. Backbone–energy encapsulates the intrinsic conformational preferences, where high-energy conformations indicate instability or unphysical conformations. \textbf{Novelty} was assessed by determining whether each plausible generated conformation is structurally distinct from its nearest neighbor in the training set. Specifically, for each generated conformation deemed plausible by the RF classifier, we applied a $k$‑nearest‑neighbors (KNN) search to retrieve its closest training conformations based on RMSD. Novel conformations indicate that MoDyGAN is able to learn the previously unsampled region of the conformational landscape.

For Ala\textsubscript{10}, the \textit{generator training dataset} only contains the initial and final conformations, leaving 18 intermediate states unseen by MoDyGAN. A generated conformation is considered \textbf{plausible} if its RMSD, Ramachandran distribution and backbone energies resemble any of the 20 reference states. It is further labeled as \textbf{novel} when these metrics most closely align with one of the 18 unseen intermediates, indicating that MoDyGAN has generalized beyond its training data. Ramachandran plots compare backbone torsion landscapes, where structurally similar conformations produce comparable distributions. We quantify their similarity by the Earth Mover’s Distance (EMD), with lower EMD indicating closer steric agreement, and Watson’s $U^{2}$ statistic for angular data. Because Ala\textsubscript{10} has relatively few dihedral samples, $U^{2}$ significance is assessed via permutation testing. A small $U^{2}$ accompanied by a large $p$‑value implies no statistically significant difference between the two angle distributions. Due to Ala\textsubscript{10}'s high flexibility, we did not train a separate RF classifier.

\subsection{Benchmarks}
\textbf{Model ablation study:} To assess the contribution of individual components in MoDyGAN, we conducted an ablation study by evaluating four model variants. \textbf{Recover} produces conformations without refinement, serving as a baseline to evaluate the impact of refiners. \textbf{MoDyGAN (Orig.)} uses one refiner trained with the original Pix2Pix architecture without the secondary structure–focused discriminator ($D_2$). \textbf{MoDyGAN (Refiner1)} and \textbf{MoDyGAN (Refiner2)} each use one refiner trained with the modified Pix2Pix model with $D_2$, applied independently to enhance generated conformations. \textbf{MoDyGAN (Ensemble)} combines Refiner1 and Refiner2 through ensemble learning to leverage their complementary strengths during refinement.

\textbf{Comparison with autoencoder model~\cite{degiacomi2019coupling}:} 
The autoencoder model from \cite{degiacomi2019coupling} is publicly available along with a well-trained model, a testing dataset comprising 200 latent vectors, and well-documented MD simulation procedures. As reference data are provided only for 1POA and 2WJ7, our comparisons are limited to these two proteins (see Supplementary Text C for details).

\section{Results and Discussion}
  
\subsection{Refiner Enhances Structural Plausibility}

\begin{figure}
\centering

\makebox[\textwidth][c]{%
    \begin{minipage}{0.18\textwidth}
        \centering \textbf{1POA}
    \end{minipage}
    \begin{minipage}{0.18\textwidth}
        \centering \textbf{2WJ7}
    \end{minipage}
    \begin{minipage}{0.18\textwidth}
        \centering \textbf{1BMR}
    \end{minipage}
    \hspace{1cm}
    \begin{minipage}{0.18\textwidth}
        \centering \textbf{Ala$_{10}$}
    \end{minipage}
}

\vspace{0.0cm}

\begin{subfigure}[b]{0.18\textwidth}
    \includegraphics[width=\linewidth]{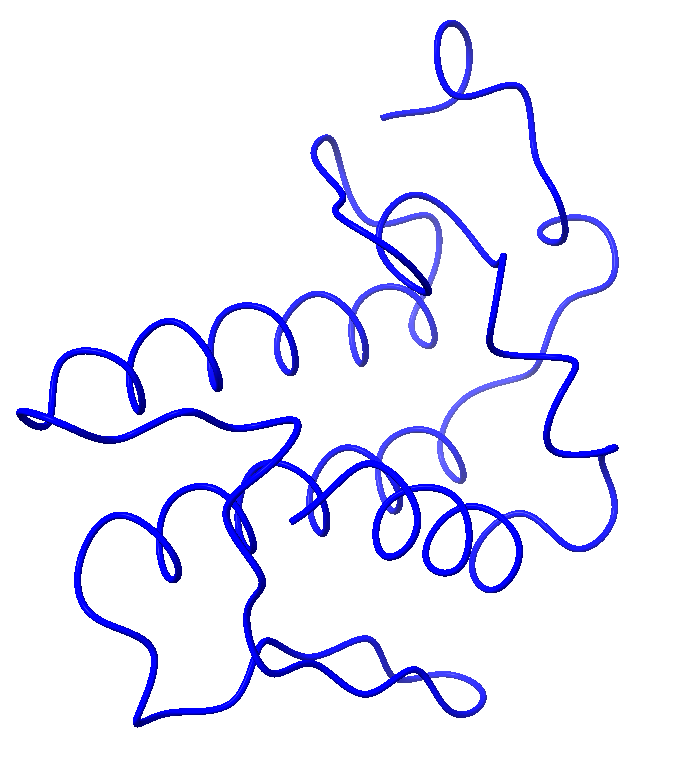}
    \centering\footnotesize Baseline \quad\quad\quad\quad(1.95 ± 0.20)
\end{subfigure}
\begin{subfigure}[b]{0.18\textwidth}
    \includegraphics[width=\linewidth,height=\linewidth]{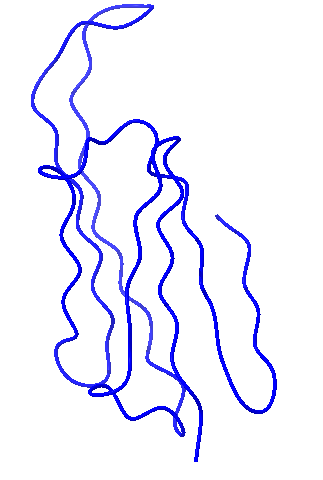}
    \centering\footnotesize Baseline \quad\quad\quad\quad(1.46 ± 0.21)
\end{subfigure}
\begin{subfigure}[b]{0.18\textwidth}
    \includegraphics[width=\linewidth]{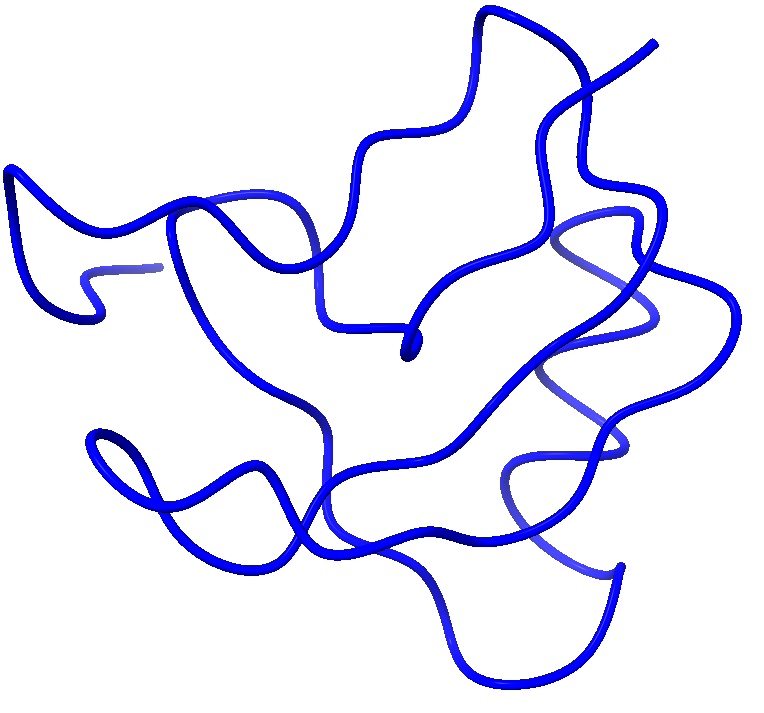}
    \centering\footnotesize Baseline \quad\quad\quad\quad(2.33 ± 0.16)
\end{subfigure}
\hspace{1cm}
\begin{subfigure}[b]{0.18\textwidth}
    \includegraphics[width=\linewidth]{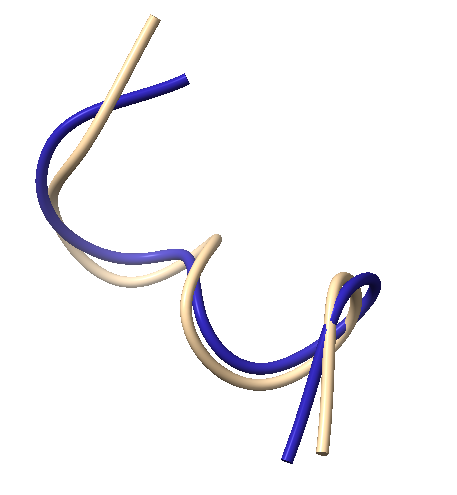}
    \centering\footnotesize Initial state\quad\quad\quad\quad (3.13 ± 0.30)
\end{subfigure}

\vspace{0.0cm}

\begin{subfigure}[b]{0.18\textwidth}
    \includegraphics[width=\linewidth]{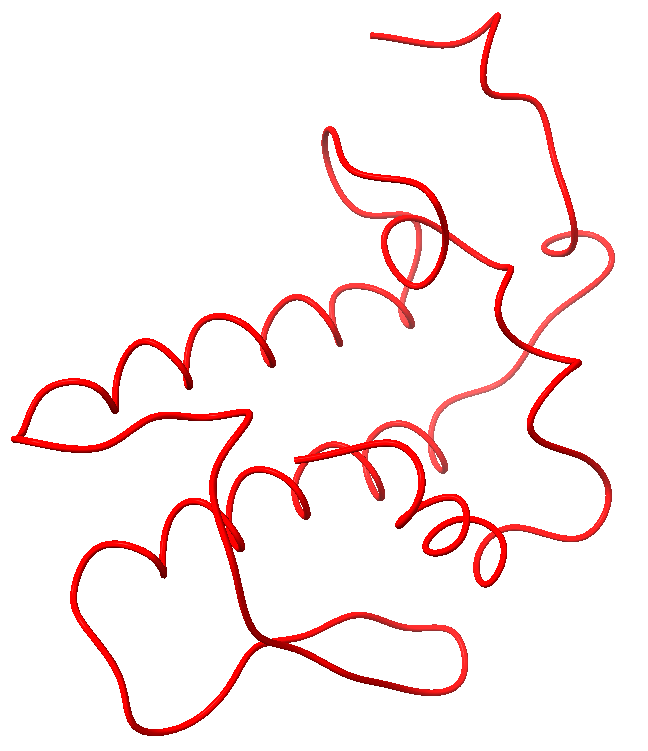}
    \centering\footnotesize Before refinement \quad\quad (2.00 ± 0.30)          
\end{subfigure}
\begin{subfigure}[b]{0.18\textwidth}
    \includegraphics[width=\linewidth,height=\linewidth]{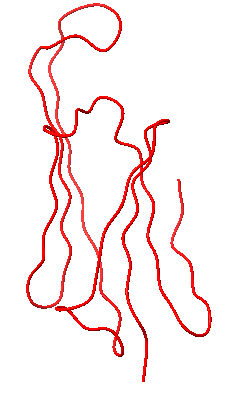}
    \centering\footnotesize Before refinement \quad\quad(1.42 ± 0.32)
\end{subfigure}
\begin{subfigure}[b]{0.18\textwidth}
    \includegraphics[width=\linewidth]{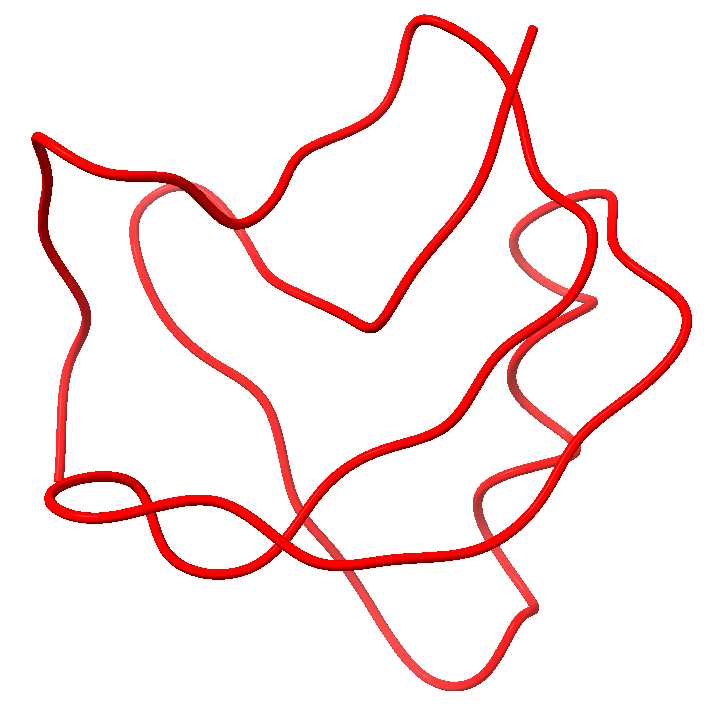}
    \centering\footnotesize Before refinement \quad\quad(2.79 ± 0.68)
    
\end{subfigure}
\hspace{1.cm}
\begin{subfigure}[b]{0.18\textwidth}
    \includegraphics[width=\linewidth]{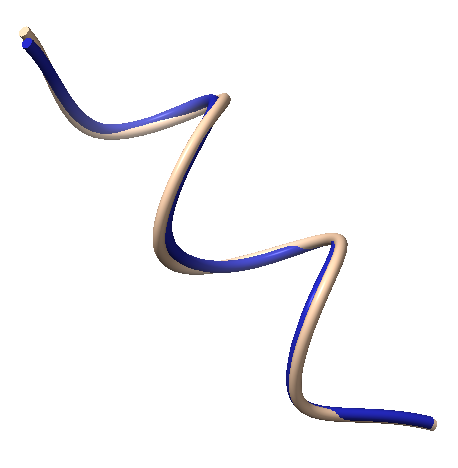}
    \centering\footnotesize State6 \quad\quad\quad\quad\quad\quad(0.88 ± 0.32)
\end{subfigure}

\vspace{0.0cm}

\begin{subfigure}[b]{0.18\textwidth}
    \includegraphics[width=\linewidth]{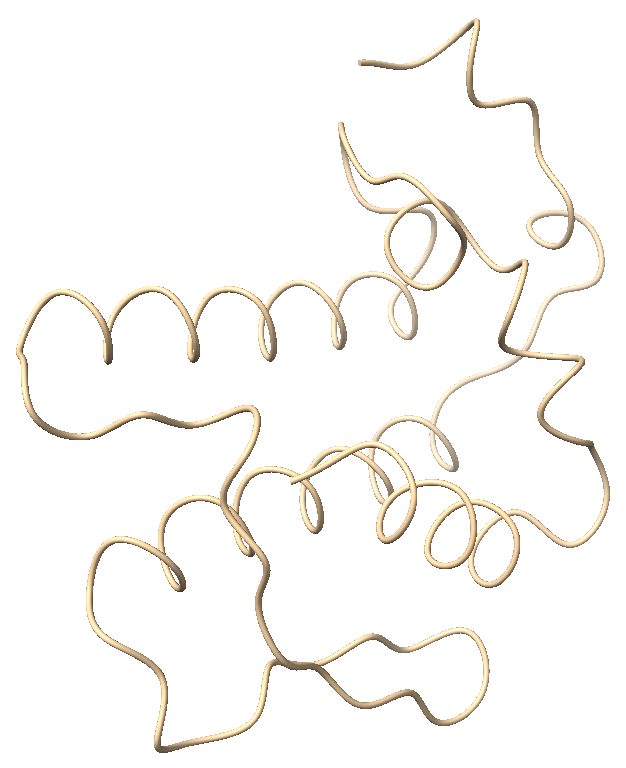}
    \centering\footnotesize After refinement \quad\quad\quad (2.00 ± 0.26)
\end{subfigure}
\begin{subfigure}[b]{0.18\textwidth}
    \includegraphics[width=\linewidth,height=\linewidth]{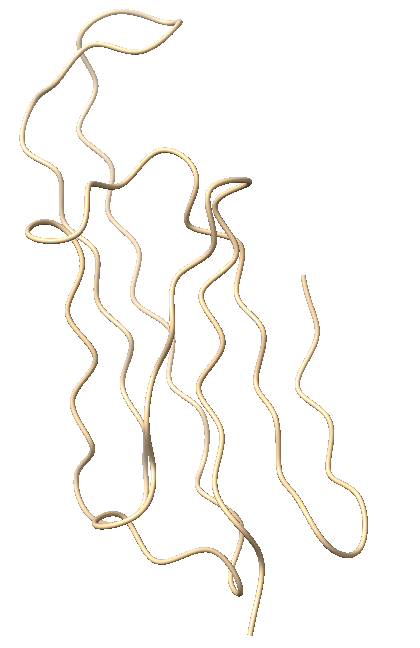}
    \centering\footnotesize After refinement \quad\quad\quad(1.31 ± 0.24)
\end{subfigure}
\begin{subfigure}[b]{0.18\textwidth}
    \includegraphics[width=\linewidth]{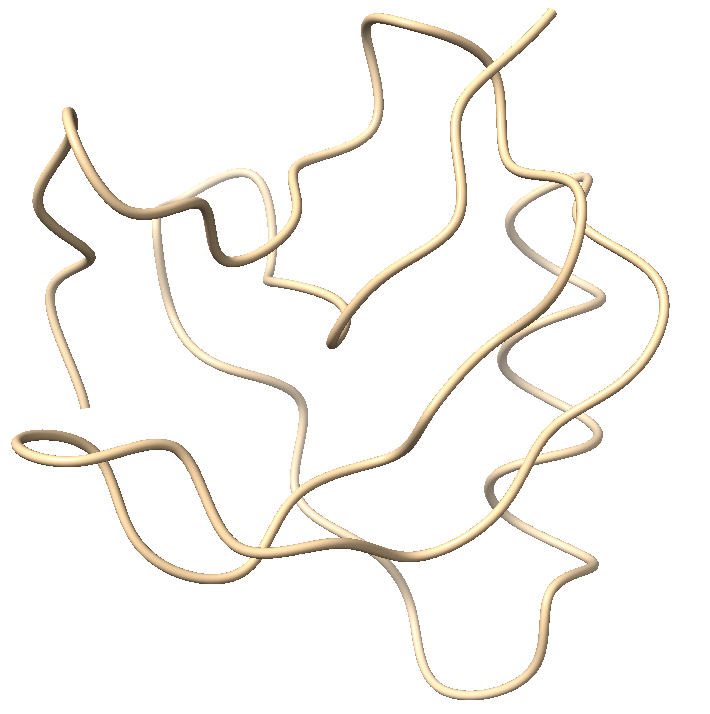}
    \centering\footnotesize After refinement \quad\quad\quad(2.48 ± 0.56)
\end{subfigure}
\hspace{1cm}
\begin{subfigure}[b]{0.18\textwidth}
    \includegraphics[width=\linewidth]{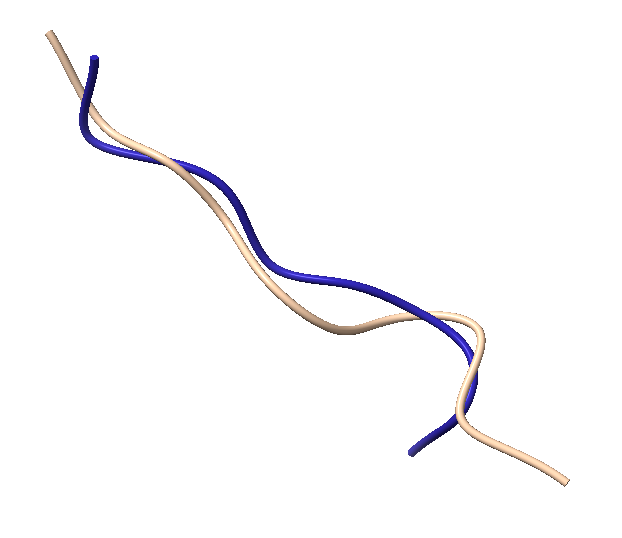}
    \centering\footnotesize Final state\quad\quad\quad\quad (2.24 ± 0.00)
\end{subfigure}

\caption{The first 3 columns show the conformations of 3 rigid proteins. 
The final column shows three representative states of Ala\textsubscript{10}, with baseline SMD structures in brown and MoDyGAN-generated conformations in blue. 
A complete comparison of 20 Ala\textsubscript{10} states is available in Supplementary Figure~S~\ref{fig:Ala_output}. RMSD is shown in parentheses as mean ± standard deviation in \AA.}
\label{fig:vis_results}
\end{figure}

Refiners significantly improve the structural quality of generator outputs by correcting structural inaccuracies, particularly in secondary structure segments, as shown in Figure~\ref{fig:vis_results}. Supplementary Table~S~\ref{tab:whole} further supports this observation. It shows refined conformations resemble the baseline structures in all metrics. Taking 1POA as an example, the refiner model improved the N--C$\alpha$--C angle from 144.92~$\pm$~13.85$^\circ$ to 115.12~$\pm$~6.21$^\circ$, which is very close to the baseline value of 112.59~$\pm$~3.84$^\circ$. The model also increased the acceptance rate for RF classification by approximately 70\%, like in 2WJ7, the acceptance rate improved from 5.12\% to 93.81\%. Additionally, it produced lower backbone energies compared to the baselines. For example, in 1BMR, the baseline energy was 126.03~$\pm$~19.86 REU, whereas the refined conformation had a significantly lower energy of 72.84~$\pm$~38.89 REU. Thus, MoDyGAN is capable of capturing a broader conformational landscape that includes thermodynamically favorable structures.

\textbf{MoDyGAN (Ensemble)} is particularly effective at generating plausible conformations for proteins enriched in secondary structure segments. Compared to \textbf{MoDyGAN (Orig.)}, it achieves comparable whole-structure performance for 1POA and 1BMR, but shows a marked improvement for 2WJ7, with over a 10\% increase in RF acceptance rate from 81.38\% to 93.81\% (Supplementary Table~S~\ref{tab:whole}). This advantage likely stems from its enhanced ability to refine and stabilize secondary structure regions. As shown in Supplementary Tables~S~\ref{tab:1poa_part}--S~\ref{tab:1bmr_part}, \textbf{MoDyGAN (Ensemble)} consistently improves RF acceptance rates within secondary structure segments. The most significant improvement was observed in the BETA6 region of 2WJ7, where the accuracy increased from 87.89\% to 97.25\%. As 2WJ7 has a higher proportion of backbone atoms located within secondary structures (63.53\%) compared to 1POA (52.54\%) and 1BMR (40.29\%), it markedly benefits from structure-aware refinement.

Our study adopted ensemble learning, combining the strengths of two top-performing refiners, to enhance structural plausibility across both secondary and flexible regions. 
It outperforms single-refiner setups in 1POA and 2WJ7, improving RF acceptance rates by 1--3\% (see Supplementary Table~S~\ref{tab:whole}). However, the improved regions are different. In 1POA, although the overall RF acceptance rate improves from 71.88 \% or 71.95\% to 74.05\%, the secondary structure accuracy remains comparable to that of a single refiner less than 2\% difference (Supplementary Table~S~\ref{tab:whole} and S~\ref{tab:1poa_part}), indicating that improvements primarily occur in non-secondary regions. While 1BMR exhibits significantly enhanced RF accuracy in secondary structures (Supplementary Table~~S~\ref{tab:1bmr_part}). For example, int BETA2 segment of 1BMR, it increased from 53.10\% or 52.77\% to 80.58\%. This highlights improvements on secondary structure segments. Since our pipeline already incorporates a secondary structure-aware discriminator, ensemble refinement provides complementary benefits by broadening correction coverage across the entire conformation.

\subsection{Exploring Novelty in Conformational Space}
MoDyGAN is capable of exploring unseen regions of the conformational landscape, thus generating novel structures rather than simply reproducing conformations from the \textit{generator training dataset}. As shown in Table~\ref{tab:knn}, nearest-neighbor analysis reveals that a substantial proportion of MoDyGAN-generated conformations, for example up to 27.3\% ($k=1$) and 97.8\% ($k=3$) for 2WJ7, have their nearest neighbors in the dataset that MoDyGAN has never seen before. In addition, these novel conformations maintain low RMSD values with their closest non-training neighbors, typically under 2~\AA. That indicates they are not only new, but also structurally similar to realistic, physically plausible MD simulation-derived conformations. For conformations closer to the training dataset, non-zero RMSD values still indicate meaningful variations rather than exact reproductions. These findings reflect MoDyGAN’s ability to generate both novel and plausible conformational landscapes, which is essential for modeling dynamic biomolecular processes.

\begin{table}[h]
\centering
\begin{threeparttable}
\caption{KNN Analysis of MoDyGAN Outputs}
\label{tab:knn}
\centering
\begin{tabular}{|l|l|l|l|l|}
\hline
\textbf{Protein} & Ratio ($k=1$)\textsuperscript{a} & Ratio ($k=3$)\textsuperscript{b} & RMSD\textsuperscript{c} (TS,\AA) & RMSD\textsuperscript{d} (Non-TS,\AA) \\ \hline
\textbf{1POA}    & 0.161 & 0.847 & 1.456 ± 0.162 & 1.504 ± 0.196 \\ \hline
\textbf{2WJ7}    & 0.273 & 0.978 & 0.890 ± 0.124 & 0.904 ± 0.109 \\ \hline
\textbf{1BMR}    & 0.332 & 0.969 & 1.743 ± 0.262 & 1.595 ± 0.208 \\ \hline
\end{tabular}

\begin{tablenotes}[flushleft]
\small
\item[] For each refined conformation generated by the \textbf{MoDyGAN (Ensemble)}, we identify its closest or top three closest matches from the MD simulation trajectory:
\item[a] Proportion of cases where the closest match is \textbf{not} in the \textit{generator training set}.
\item[b] Proportion of cases where at least one of the top three closest matches are \textbf{not} in the \textit{generator training set}.
\item[c] For cases in (a), RMSD (mean ± std) to the nearest match \textbf{within} the training set.
\item[d] For cases in (a), RMSD (mean ± std) to the nearest match \textbf{outside} the training set.
\end{tablenotes}
\end{threeparttable}
\end{table}

\subsection{Learning Flexible Protein Conformational Space}

For flexible proteins, MoDyGAN can partially learn the unseen conformational landscape, especially in helical regions, but struggles with highly dynamic or disordered states. As shown in Figure~\ref{fig:vis_results} and Supplementary Figure~S~\ref{fig:Ala_output}, MoDyGAN-generated structures align well with SMD references across most states, particularly those with pronounced helical characteristics (e.g., states 0–10). In contrast, later states (after 15) with greater variability show mismatches. Further validation is provided in Supplementary Table~S~\ref{tab:10ala_backbone_comparison}. When evaluated using RMSD alone, MoDyGAN-generated conformations showed varying degrees of structural similarity across intermediate reference states, with states 5 to 11 exhibiting particularly low average RMSDs (below 2~\AA). Adding dihedral constraints (\(U^2 < 0.2\), \(p > 0.05\)) reduced the RMSD. However, several reference states, especially states 0 to 6, state 8 and states 13 to 15, still retain over 10 conformations with EMD ranges from 0.02 to 0.05. This suggests that MoDyGAN can produce new conformations that are both spatially and torsionally plausible, particularly in well-structured helical regions.


However, these conformations often exhibit higher backbone energies than reference states, indicating physical instability and the need for energy-aware refinement. As shown in Supplementary Table~S~\ref{tab:10ala_backbone_comparison}, applying an energy constraint, which requires lower energy than the corresponding baseline, limits valid outputs to five states. They satisfy all three plausible criteria: low RMSD, plausible dihedral angles, and favorable backbone energy. Especially, conformations close to state5 and state6 achieve significantly lower energies, with \(10.499 \pm 3.956\)~REU and \(0.607 \pm 4.179\)~REU, respectively. As these states predominantly exhibit helical features, the results support our earlier visualization findings, highlighting that MoDyGAN performs particularly well in modeling structured and energetically favorable regions, such as stable helices. The results suggest that energy-aware refinement may be needed to further improve spatially plausible yet energetically unstable conformations.

\subsection{Bridging Latent Space and the Conformational Landscape}

MoDyGAN learns a smooth and meaningful latent space that supports continuous transitions between valid conformations. For comparison, in \cite{degiacomi2019coupling}, not every coordinate in latent space can be decoded into a physically plausible conformation. In contrast, this work shows that MoDyGAN maps samples randomly from a Gaussian distribution onto conformations with high RF acceptance rates. This demonstrates MoDyGAN effectively bridges latent space coordinates to physically valid molecular conformations.
 Furthermore, as illustrated in Figure~\ref{fig:vis_results} and Supplementary Figure~S~\ref{fig:Ala_output}, interpolations between initial and final states yield gradual, physically plausible structural transformations. This highlights MoDyGAN’s ability to capture conformational dynamics continuously and realistically.

\section{Conclusion}
In this paper we presented a novel GANs-based pipeline designed to efficiently explore protein conformational spaces. We introduced an innovative representation method that is able to convert 3D protein conformations into 2D matrices reversibly, thus enabling the utilization of multiple state-of-the-art image-based GANs. Our results demonstrate that a well-trained generator can map Gaussian distributions onto realistic MD trajectories. Additionally, we present a novel refinement module employing ensemble learning coupled with a dual-discriminator framework. This module significantly enhances the quality of generated protein conformations. The combined use of generator and refiner provides a promising mechanism for exploring protein conformational landscapes, producing novel and biologically plausible conformations. Using Ala$_{10}$ as an example, we also show that interpolations within the latent space correspond closely to trajectories derived from SMD simulations.
Our method presents several limitations. These include local structural inaccuracies, particularly near flexible protein regions, limited testing across only four protein systems, and potential areas for improvement through energy-aware refinement. Despite these limitations, our work suggests proteins can be treated as \emph{images}, thereby enabling the application of advanced GAN architectures. Furthermore, the proposed representation method is not limited to proteins alone and could generalize to other complex 3D structures. The performance of our approach in generating secondary structures allows potential extensions in drug design or bioengineering, such as protein loop modeling. Future work will extend this pipeline to predict future conformations given a dynamic trajectory. 
In addition, future efforts will focus on addressing limitations, such as enhancing local structural accuracy in flexible regions, or extending validation to a wider range of proteins.

\bibliographystyle{unsrt}  
\bibliography{references.bib}

\clearpage
\appendix

\begin{titlepage}
  \centering
  {\Huge\bfseries Supplementary Material}\\[2em]
  {\Large\itshape MoDyGAN: Combining Molecular Dynamics With GANs to Investigate Protein Conformational Space}\\[1.5em]
  {\large
    Jingbo Liang\textsuperscript{1},\quad
    Bruna Jacobson\textsuperscript{1*}
  }\\[1em]
  {\small
    \textsuperscript{1}Department of Computer Science, University of New Mexico, Albuquerque, USA
  }\\[2em]
  {\footnotesize
    \textsuperscript{*}\textbf{Corresponding author:}
    \href{mailto:bjacobson@unm.edu}{bjacobson@unm.edu}
  }
  \vfill
\end{titlepage}

\setcounter{figure}{0}  
\setcounter{table}{0} 

\title{}
\newpage
\section{Proteins}
Proteins are macromolecules composed of amino acid chains. Their function and interaction are directly influenced by their 3D conformation. The protein backbone is particularly critical, as it serves as the structural framework that shapes and stabilizes these conformations.

The protein backbone is a repetitive structure composed of alternating peptide bonds and C\textsubscript{$\alpha$}. It imparts the necessary flexibility for proteins to adopt secondary structures, including $\alpha$-helices and $\beta$-sheets. Another crucial component within the backbone is their dihedral angles, $\phi$ and $\psi$. They define sterically favorable conformations that guide protein folding and enable the formation of specific tertiary and quaternary structures essential for biological function.

In our study, we include three rigid proteins and one flexible protein to represent systems with stable or dynamic secondary structures, respectively. For relatively rigid proteins, we selected: \textbf{phospholipase A\textsubscript{2}} (PDB ID: 1POA), consisting of 354 backbone atoms and characterized by three $\alpha$-helices and a short $\beta$-strand; \textbf{chain A of $\alpha$B-crystallin} (PDB ID: 2WJ7), consisting of 255 backbone atoms and characterized by a $\beta$-strand–dominated architecture; and the \textbf{alpha-like toxin LqH III} (PDB ID: 1BMR), consisting of 201 backbone atoms and characterized by a short $\alpha$-helix along with a three-stranded antiparallel $\beta$-sheet. For highly flexible proteins, we selected \textbf{deca-alanine} (Ala\textsubscript{10}), a synthetic polypeptide composed of ten alanine residues. Ala\textsubscript{10} experiences significantly helix–coil transition during SMD simulation, where the C-terminal C${\alpha}$ is pulled while the N-terminal C${\alpha}$ remains fixed under applied force (see Supplementary Figure ~S~\ref{supp:ala10}).

\newpage
\section{Molecular Dynamics Simulations}

\begin{figure}[H]
    \renewcommand{\figurename}{Figure S\hskip-\the\fontdimen2\font\space }
    \centering    
    \includegraphics[width=1\linewidth]{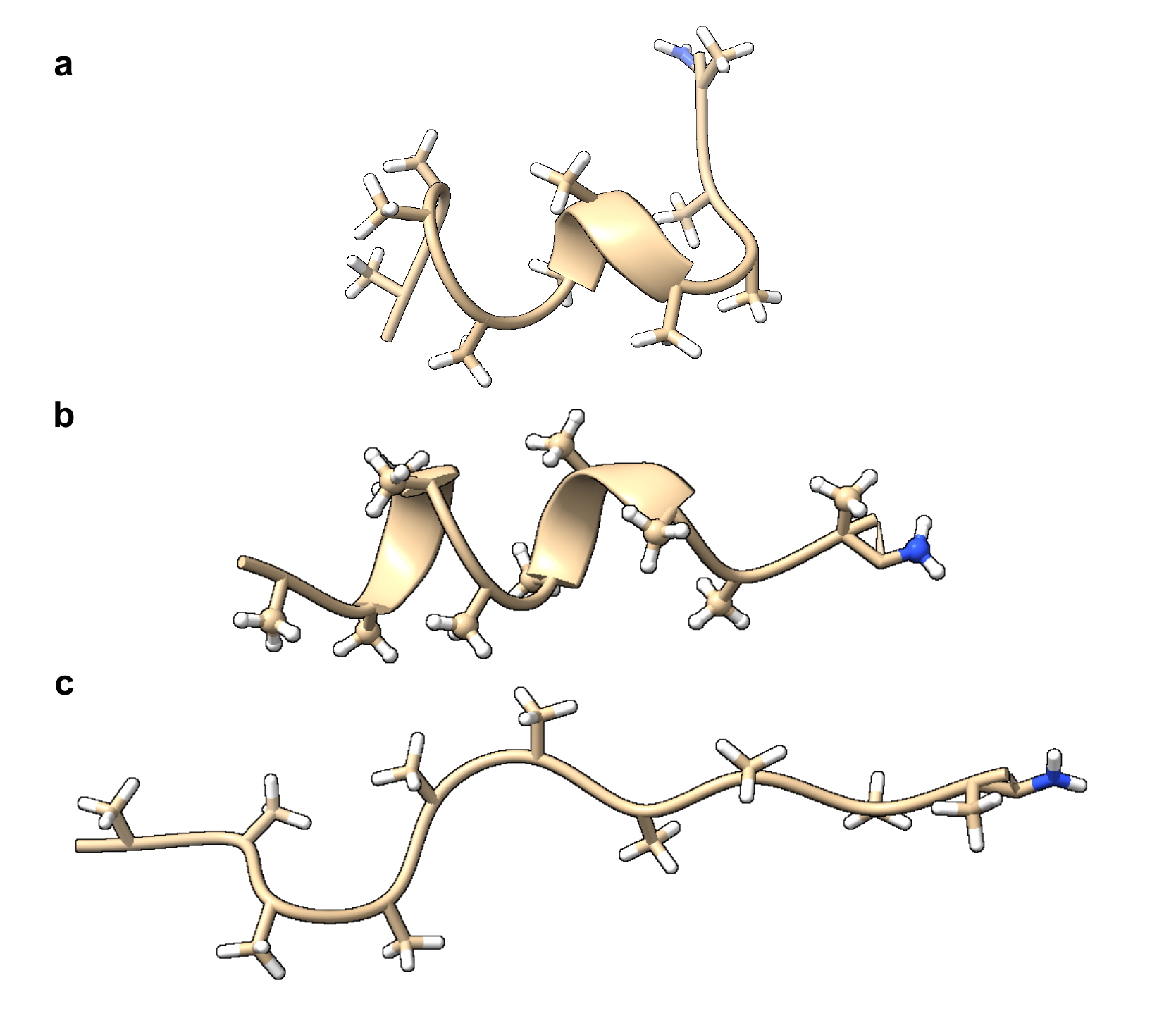}
    \caption{Three snapshots of unfolding Ala$_{10}$ at different stages of the SMD simulation. (a) shows the initial crystal structure. A folded intermediate state (b) forms as the helix begins to unravel at the terminal. The structure fully unfolds into an extended state (c).}
    \label{supp:ala10}
\end{figure}

Steered molecular dynamics (SMD) of Ala$_{10}$ were performed using the 104-atom compact helical model from [1], following detailed procedural steps from [2]. SMD eventually output 20 different conformational states. We kept both the initial (Supplementary Figure~S~\ref{supp:ala10}a) and final states (Supplementary Figure~S~\ref{supp:ala10}c) as the starting point for the following MD simulations. The other 18 conformations were kept as reference states.

MD simulations were conducted on protein 1POA, 2WJ7, 1BMR, the initial and final states of Ala$_{10}$. MD simulations utilized the CHARMM27 force field implemented within the NAMD software [3]. Each system was solvated with TIP3P water molecules and neutralized with Na$^{+}$ and Cl$^{-}$ ions within a spherical environment. We applied the simulation at a temperature of 310~K maintained by Langevin dynamics. A timestep of 2 fs was employed with all bonds constrained for numerical stability. Non-bonded interactions employed a 12 Å cutoff with a switching function initiated at 10 Å and a pair-list distance of 14 Å. Electrostatics were calculated using PME, with a grid spacing of 2.5 Å and periodic boundary conditions. Simulations output included restart files every 500 steps (1 ps), trajectory data every 250 steps (0.5 ps), and energy information every 100 steps (0.2 ps). The simulations were executed for a duration exceeding 0.5 ns.

\newpage
\section{Supplementary Evaluation}
\subsection{Random Forest (RF) classification}
Assessing if a structure is a physically plausible protein conformation can be formulated as a binary‑classification problem.  A conformation with \(N\) atoms is represented by a single point in a \(3 \times N\)-dimensional coordinate space. RF classification can partition this space into two regions: one corresponding to plausible protein geometries and the other to implausible ones. We only used RF classification assessment on three rigid proteins.

We first created two datasets. \textit{Random forest (RF) dataset} includes 20{,}000 conformations, divided into two classes: \textit{correct} structures, directly obtained from MD simulations, and \textit{perturbed} structures, generated by adding 0.5~\AA\ random noise to atomic coordinates. \textit{Validation dataset} includes all the remaining conformations not used in the other sets. For each protein, only backbone atoms were considered.

We then instantiated a RF classifier and defined a hyperparameter search space (e.g., \texttt{n\_estimators}, \texttt{max\_depth}, and so on). Using \texttt{RandomizedSearchCV} from \texttt{scikit‑learn}, we sampled a predefined number of hyperparameter combinations and selected the best estimator. The resulting best estimator was subsequently evaluated on the validation set.

\subsection{Root‑mean‑square deviation (RMSD)}
Root‑mean‑square deviation (RMSD) quantifies the average atomic difference between two molecular structures, thus providing an intuitive measure of their structural similarity. In this study we compute RMSD over backbone atoms in two superimposed molecular structures, \(M\) and \(M'\).RMSD is defined in Equation ~\eqref{eq:rmsd},where \(N\) is the number of backbone atoms, and \(\mathbf{r}_{i}\) and \(\mathbf{r}'_{i}\) denote the Cartesian coordinates of the \(i^{\text{th}}\) atom in \(M\) and \(M'\).

\begin{equation}
\text{RMSD} \;=\;
\sqrt{\frac{1}{N}\sum_{i=1}^{N} \bigl\lVert \mathbf{r}_{i}-\mathbf{r}'_{i}\bigr\rVert^{2}}
\label{eq:rmsd}
\end{equation}

\subsection{Bond and Angle Analysis}
In this study we evaluate bond lengths (C–C$_\alpha$, N–C, C$_\alpha$–N), the backbone bond angle (N–C$_\alpha$–C), and the dihedral angles $\phi$ and $\psi$—as summarised in Supplementary Table S~\ref{tab:s2}.
 
\subsection{KNN Algorithm}
Assessing if a structure is new can be formulated as \(k\)-nearest‑neighbors (KNN) problem. A conformation with \(N\) atoms is represented by a single point in a \(3 \times N\)-dimensional coordinate space. Given a generated structure, the KNN algorithm identifies its closest neighbors among the sampled structural clusters.

Using \texttt{scikit‑learn}'s \texttt{KNeighborsClassifier}, we fit a KNN model on all samples from the MD simulation and employ it to assign each MoDyGAN‑generated structure its nearest neighbors. If one's nearest neighbors is not within \textit{generator training dataset}, then this conformation is considered as new.

\subsection{Backbone Energy}
Backbone energy is calculated based on the protein's main‑chain dihedral angles and peptide-bond planarity. It serves as an indicator of structural plausibility. Because a protein with unfavourable backbone energy is unstable. 

In our study, we first processed each structure with \texttt{PDBFixer} [4].\texttt{PDBFixer} is in charge of adding any missing heavy atoms, rebuilding side chains and protonating the model at~pH~7.0 without changing the coordinates of atoms already present in the input PDB file. Finally, the resulting repaired structure is loaded into \texttt{PyRosetta} [1] to compute the backbone energy only.

\subsection{Ramachandran Plots, EMD and Watson’s U² statistic}
The Ramachandran plot displays the backbone dihedral angles for each residue in a protein, delineating sterically allowed and forbidden regions. Hence, two proteins that share nearly identical Ramachandran plots possess essentially the same dihedral distributions, indicating closely matched backbone conformations, especially for the secondary structure part.

To evaluate the similarity between Ramachandran plots, we used two metrics. 

\textbf{Earth Mover’s Distance (EMD):}  A Ramachandran plot could be converted into a 2D histogram. We treat this histogram as a pile of \emph{earth}, where the height of every bin represents the amount of mass in that dihedral region. The EMD quantifies the minimal work required to reshape one pile into the other. A large EMD indicates markedly different backbone‑torsion distributions, whereas a small value indicates similar distributions.

In our paper, Ramachandran plots generated from 20 intermediate states of $Ala_{10}$ are the baseline. We also generated Ramachandran plots for our generated conformations and constructed 2D histograms for both. These histograms were flattened into 1D vectors of bin counts, and Wasserstein distances were computed using \texttt{scipy.stats.wasserstein\_distance}.

\textbf{Watson’s U² test(\(\phi\) or \(\psi\))):} Watson’s \(U^{2}\) statistic provides a non‑parametric test for quantifying the difference between two angular distributions. Larger \(U^{2}\) values indicate substantial divergence, whereas smaller ones imply similarity.  Because Ala\textsubscript{10} supplies only 9 dihedral angles of either phi or psi, we assess significance via permutation testing. Specifically,  we first extracted the \(\phi\) or \(\psi\) angles from the two conformations under comparison, arranged each set as a 1D vector, and computed the observed \(U^{2}\). The two angle sets were then pooled, randomly shuffled, and repartitioned into groups of nine for 10\,000 permutations. The \(p\)-value is the fraction of permuted scores that equal or exceed the observed statistic. In this study, we consider a generated conformation’s torsion distribution to match the reference when \(U^{2} < 0.2\) and \(p > 0.05\), indicating no statistically significant difference.

\subsection{Autoencoder Model Evaluation}
We sampled 5{,}040 conformations for 1POA and 2WJ7 each by following the provided MD simulation protocols~[6]. We used only chain A of 2WJ7 in our experiments, whereas the decoder generated multi-chain structures. To ensure a fair comparison, we isolated chain A from the decoder's output. They served as the ground-truth sets. The first conformation from the MD trajectory serving as the RMSD reference structure. Then, using the provided trained autoencoder and the testing dataset, we generated 200 conformations each. Bond/angle analysis and backbone energy evaluation were performed between the generated conformations and the ground-truth set. RMSD was performed between the generated conformations and the reference structure. The generation process requires a scaling factor to restore generated conformations. However, the exact scaling factor was not provided by the authors. Thus, we empirically selected the value that minimized RMSD between generated and ground-truth structures.

\newpage
\section{Supplementary Conformation Representation Recovery}
We first denormalize and remove padding from the generated matrices. The $i$-th row in the resulting matrix describes the conformation in spherical coordinates centered at atom $i$. Let $\mathbf{S}_i \in \mathbb{R}^{n \times 3}$ denote the spherical coordinate matrix derived from the $i$-th row~(Equation~\ref{eq:s}). We then convert $\mathbf{S}_i$ to $\mathbf{T}_i \in \mathbb{R}^{n \times 3}$, which is the corresponding Cartesian coordinates (see Equation~\ref{eq:vector_spherical_to_cartesian}).

\begin{equation}
\mathbf{S}_i =
\begin{bmatrix}
r_1 & \phi_1 & \theta_1 \\
r_2 & \phi_2 & \theta_2 \\
\vdots & \vdots & \vdots \\
r_n & \phi_n & \theta_n
\end{bmatrix}
\in \mathbb{R}^{n \times 3}
\label{eq:s}
\end{equation}

\begin{equation}
\mathbf{T}_i =
\begin{bmatrix}
x_1 & y_1 & z_1 \\
x_2 & y_2 & z_2 \\
\vdots & \vdots & \vdots \\
x_n & y_n & z_n
\end{bmatrix}
=
\begin{bmatrix}
r_1 \sin(\phi_1) \cos(\theta_1) & r_1 \sin(\phi_1) \sin(\theta_1) & r_1 \cos(\phi_1) \\
r_2 \sin(\phi_2) \cos(\theta_2) & r_2 \sin(\phi_2) \sin(\theta_2) & r_2 \cos(\phi_2) \\
\vdots & \vdots & \vdots \\
r_n \sin(\phi_n) \cos(\theta_n) & r_n \sin(\phi_n) \sin(\theta_n) & r_n \cos(\phi_n)
\end{bmatrix}
\in \mathbb{R}^{n \times 3}
\label{eq:vector_spherical_to_cartesian}
\end{equation}

\newpage
\section{Supplementary Figures}

\begin{figure}[H]
    \renewcommand{\figurename}{Figure S\hskip-\the\fontdimen2\font\space }
    \centering      
    \includegraphics[width=0.7\linewidth]{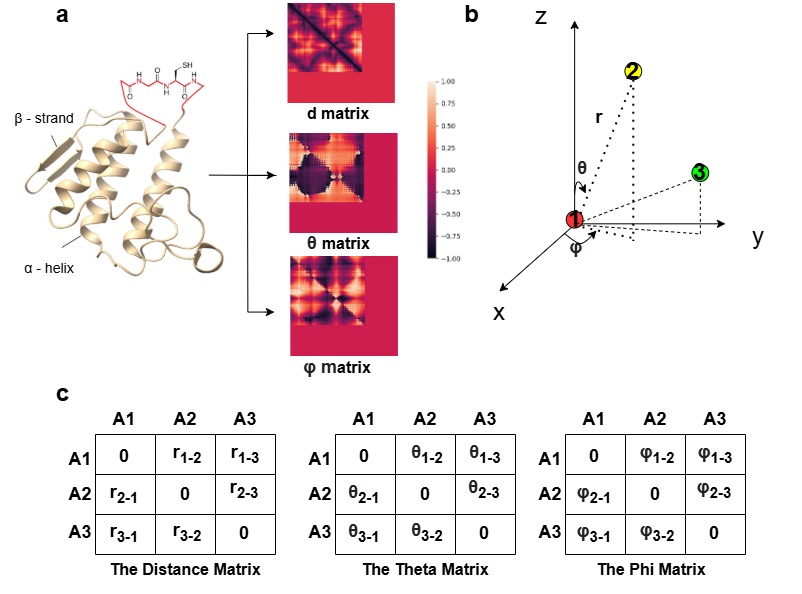}
    \caption{Protein pairwise feature representation. (a) Using 1POA as an example, protein conformations are represented using pairwise feature matrices computed based on backbone atoms, including nitrogen (N), alpha carbon (C${\alpha}$), and carbon (C). (b) illustrates a simplified, pseudo-protein conformation with three atoms. (c) shows the corresponding pairwise feature matrices derived from (b). Each atom $i$ is treated as the center of a spherical coordinate system. For each other atom $j$, the Euclidean distance $r_{i-j}$, polar angle $\theta_{i-j}$, and azimuthal angle $\phi_{i-j}$ are calculated. These values populate the matrices, with each row corresponding to a different origin atom $i$ and its spatial relationships to all other atoms $j$.}
    \label{fig:proteinRepresentation}
\end{figure}

\newpage

\clearpage
\begin{figure}[htbp]
\renewcommand{\figurename}{Figure S\hskip-\the\fontdimen2\font\space }
  \centering
  \scalebox{0.8}{
  \begin{minipage}{\textwidth}
  
    \begin{minipage}[b]{0.17\textwidth}
      \centering
      \includegraphics[width=\linewidth]{figures/0.png}
      {\footnotesize State 0 (Initial state)}
    \end{minipage}
    \begin{minipage}[b]{0.17\textwidth}
      \centering
      \includegraphics[width=\linewidth]{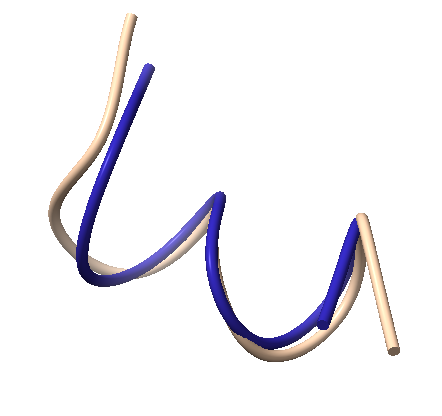}
      {\footnotesize State 1}
    \end{minipage}
    \begin{minipage}[b]{0.17\textwidth}
      \centering
      \includegraphics[width=\linewidth]{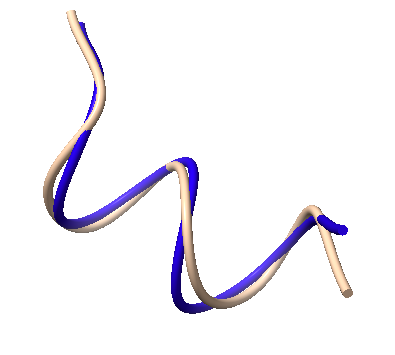}
      {\footnotesize State 2}
    \end{minipage}
    \begin{minipage}[b]{0.17\textwidth}
      \centering
      \includegraphics[width=\linewidth]{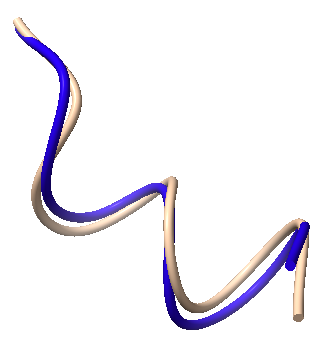}
      {\footnotesize State 3}
    \end{minipage}
    \begin{minipage}[b]{0.17\textwidth}
      \centering
      \includegraphics[width=\linewidth]{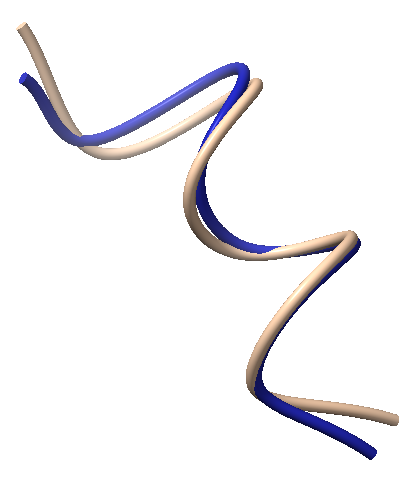}
      {\footnotesize State 4}
    \end{minipage}
  
    \vspace{8pt}
  
    \begin{minipage}[b]{0.17\textwidth}
      \centering
      \includegraphics[width=\linewidth]{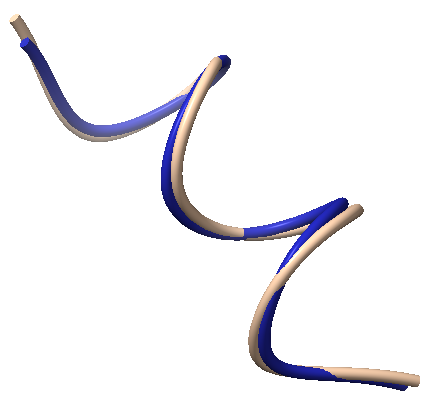}
      {\footnotesize State 5}
    \end{minipage}
    \begin{minipage}[b]{0.17\textwidth}
      \centering
      \includegraphics[width=\linewidth]{figures/6.png}
      {\footnotesize State 6}
    \end{minipage}
    \begin{minipage}[b]{0.17\textwidth}
      \centering
      \includegraphics[width=\linewidth]{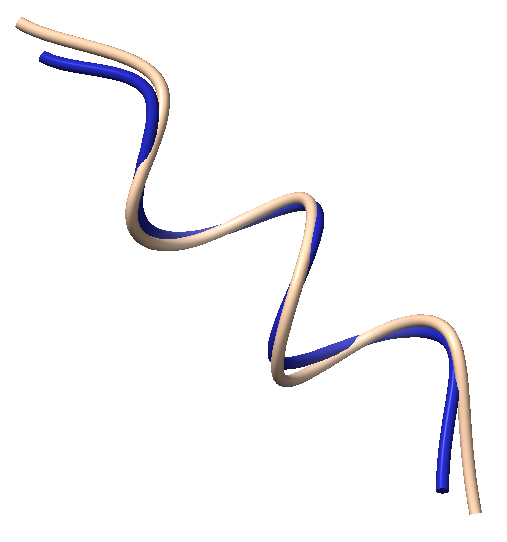}
      {\footnotesize State 7}
    \end{minipage}
    \begin{minipage}[b]{0.17\textwidth}
      \centering
      \includegraphics[width=\linewidth]{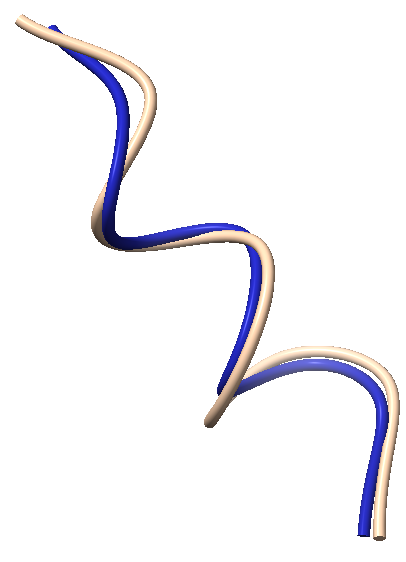}
      {\footnotesize State 8}
    \end{minipage}
    \begin{minipage}[b]{0.17\textwidth}
      \centering
      \includegraphics[width=\linewidth]{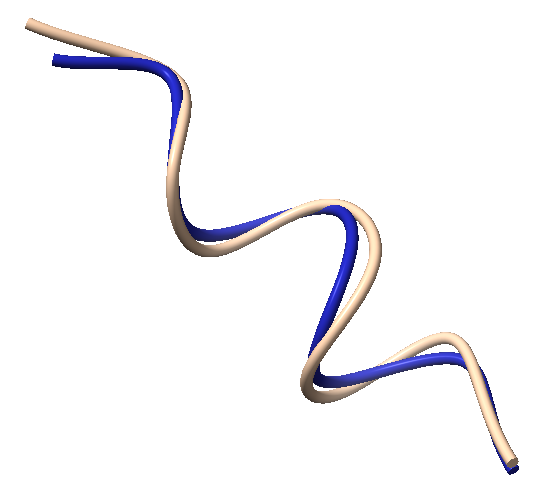}
      {\footnotesize State 9}
    \end{minipage}
  
    \vspace{8pt}
  
    \begin{minipage}[b]{0.17\textwidth}
      \centering
      \includegraphics[width=\linewidth]{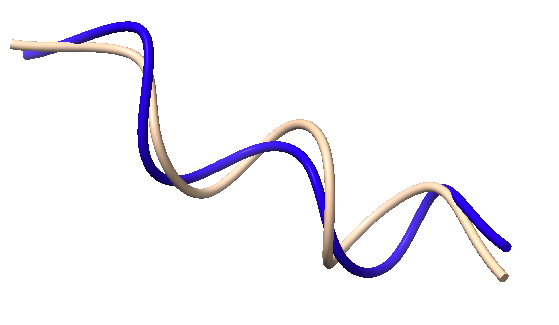}
      {\footnotesize State 10}
    \end{minipage}
    \begin{minipage}[b]{0.17\textwidth}
      \centering
      \includegraphics[width=\linewidth]{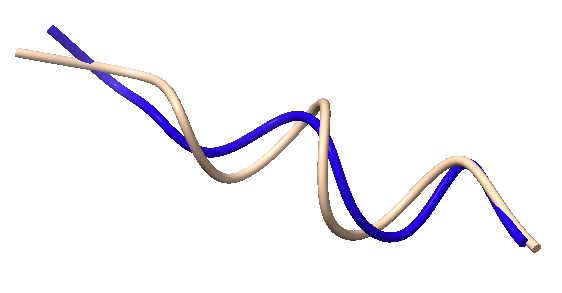}
      {\footnotesize State 11}
    \end{minipage}
    \begin{minipage}[b]{0.17\textwidth}
      \centering
      \includegraphics[width=\linewidth]{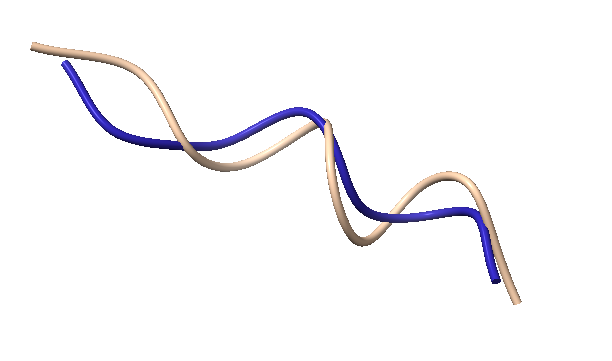}
      {\footnotesize State 12}
    \end{minipage}
    \begin{minipage}[b]{0.17\textwidth}
      \centering
      \includegraphics[width=\linewidth]{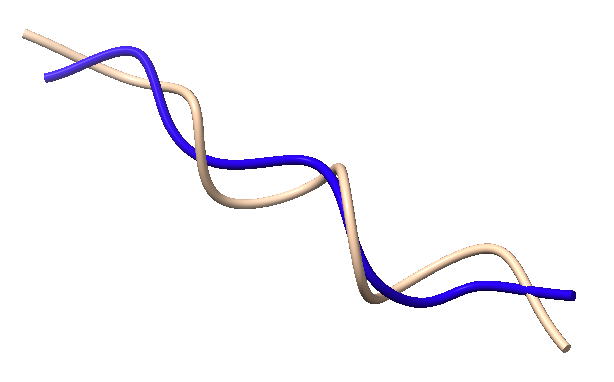}
      {\footnotesize State 13}
    \end{minipage}
    \begin{minipage}[b]{0.17\textwidth}
      \centering
      \includegraphics[width=\linewidth]{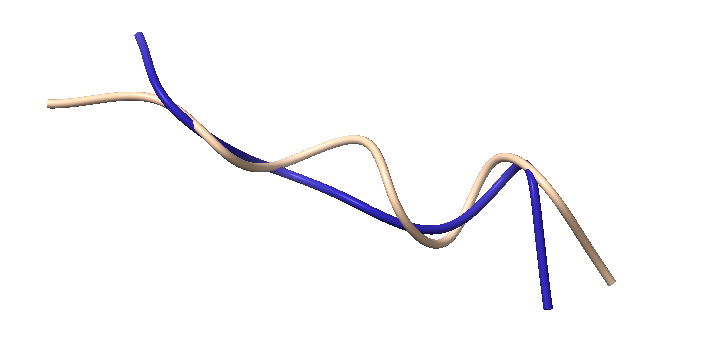}
      {\footnotesize State 14}
    \end{minipage}
  
    \vspace{8pt}
  
    \begin{minipage}[b]{0.17\textwidth}
      \centering
      \includegraphics[width=\linewidth]{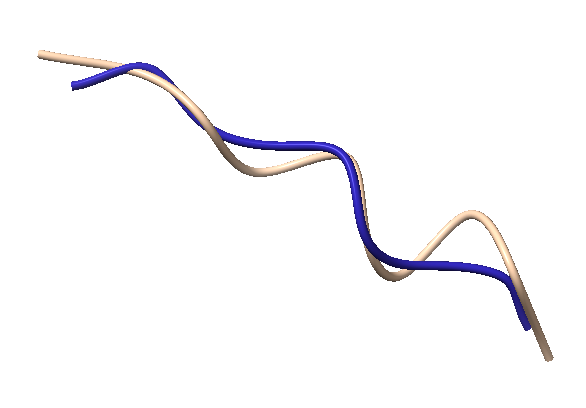}
      {\footnotesize State 15}
    \end{minipage}
    \begin{minipage}[b]{0.17\textwidth}
      \centering
      \includegraphics[width=\linewidth]{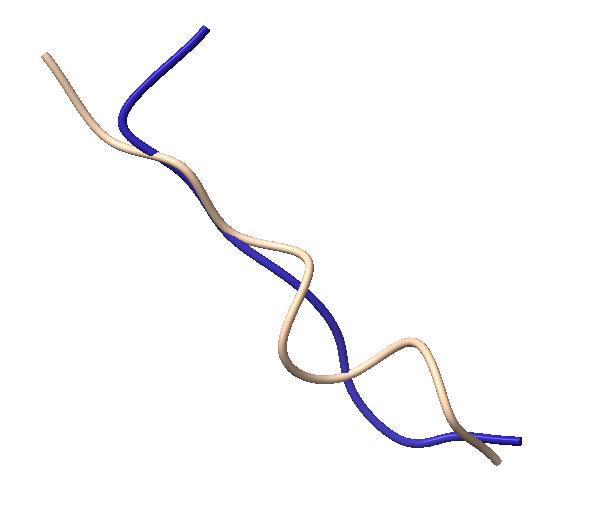}
      {\footnotesize State 16}
    \end{minipage}
    \begin{minipage}[b]{0.17\textwidth}
      \centering
      \includegraphics[width=\linewidth]{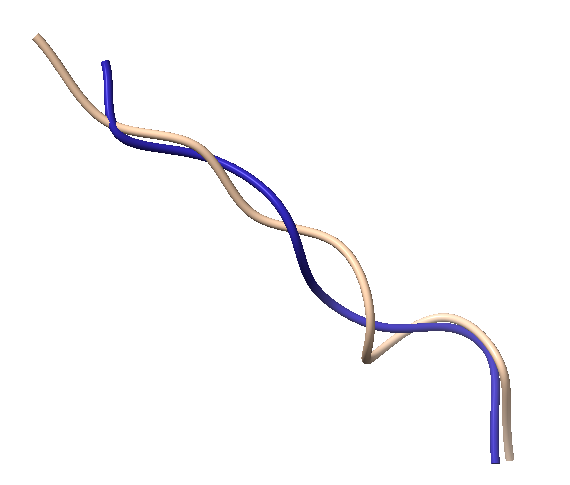}
      {\footnotesize State 17}
    \end{minipage}
    \begin{minipage}[b]{0.17\textwidth}
      \centering
      \includegraphics[width=\linewidth]{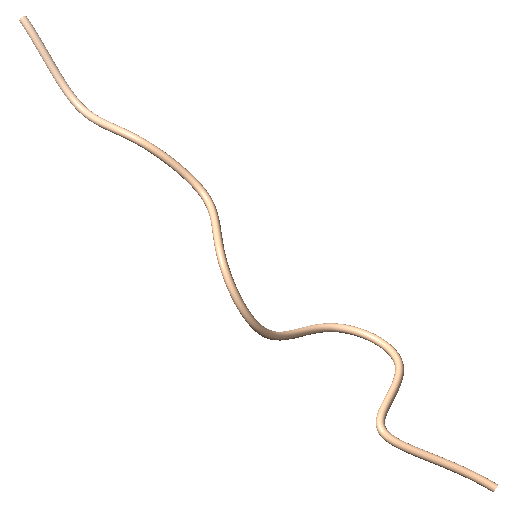}
      {\footnotesize State 18}
    \end{minipage}
    \begin{minipage}[b]{0.17\textwidth}
      \centering
      \includegraphics[width=\linewidth]{figures/19.png}
      {\footnotesize State 19 (Final state)}
    \end{minipage}
  
  \end{minipage}
  } 
  \caption{20 states of Ala\textsubscript{10}. Baseline SMD structures are shown in brown and MoDyGAN-generated conformations in blue.}
  \label{fig:Ala_output}
\end{figure}

\newpage
\section{Supplementary Tables}

\begin{table}[H] 
\centering
\renewcommand{\tablename}{Table S\hskip-\the\fontdimen2\font\space }
\caption{Training Configuration Summary for Generator and Refiner}
\label{tab:conf}
\makebox[\textwidth]{
\resizebox{1.20\textwidth}{!}{  
    
\begin{tabular}{|l|l|ll|ll|l|l|l|}
\hline
\multicolumn{1}{|c|}{\multirow{2}{*}{}} & \multicolumn{1}{c|}{\multirow{2}{*}{Structure Info\textsuperscript{a}}} & \multicolumn{2}{c|}{Generator Training\textsuperscript{b}}                       & \multicolumn{2}{c|}{Refiner Training}                        & \multicolumn{1}{c|}{\multirow{2}{*}{\begin{tabular}[c]{@{}c@{}}Best Generator\\ Epoch\textsuperscript{c}\end{tabular}}} & \multicolumn{1}{c|}{\multirow{2}{*}{\begin{tabular}[c]{@{}c@{}}Best Orig \\ pix2pix Epoch\textsuperscript{d}\end{tabular}}} & \multicolumn{1}{c|}{\multirow{2}{*}{\begin{tabular}[c]{@{}c@{}}Best Modified\\ Refiners Epoch\textsuperscript{e}\end{tabular}}} \\ \cline{3-6}
\multicolumn{1}{|c|}{}                  & \multicolumn{1}{c|}{}                                                   & \multicolumn{1}{c|}{Epoch}                     & \multicolumn{1}{c|}{Batch Size} & \multicolumn{1}{c|}{Epoch} & \multicolumn{1}{c|}{Batch Size} & \multicolumn{1}{c|}{}                                                                                                   & \multicolumn{1}{c|}{}                                                                                                       & \multicolumn{1}{c|}{}                                                                                                           \\ \hline
1POA                                    & 354                                                                     & \multicolumn{1}{l|}{[20,40,60,80,80,80,80,80]} & [32, 32, 32, 16, 16,10,4,3]     & \multicolumn{1}{l|}{50}    & 100                             & 80                                                                                                                      & 24                                                                                                                          & 25/26                                                                                                                           \\ \hline
2WJ7                                    & 255                                                                     & \multicolumn{1}{l|}{[20,40,60,80,100,100,100]} & [32, 32, 32, 16, 16, 10, 4]     & \multicolumn{1}{l|}{50}    & 100                             & 65                                                                                                                      & 36                                                                                                                          & 48/49                                                                                                                           \\ \hline
1BMR                                    & 201                                                                     & \multicolumn{1}{l|}{[20,40,60,80,100,100,100]} & [32, 32, 32, 16, 16, 10, 4]     & \multicolumn{1}{l|}{200}   & 100                             & 100                                                                                                                     & 172                                                                                                                         & 167/169                                                                                                                         \\ \hline
Ala\textsubscript{10}                   & 30                                                                      & \multicolumn{1}{l|}{[100,100,100]}             & [32, 32, 32]                    & \multicolumn{1}{l|}{240}   & 100                             & 100                                                                                                                     & 220                                                                                                                & $\backslash$                                                                                                                             \\ \hline
\end{tabular}
}
}
\end{table}
\begin{tablenotes}[flushleft]
      \small
      \item{a} Number of atoms in backbone.
      \item{b} The generator uses progressive training strategy, thus configurations vary by step. Each entry in the list denotes the epoch count or the batch size used at the $i$-th progressive step.
      \item{c} Generator was selected based on its performance at the specified epoch in the last step of progressive training.  
      \item{d} Refiner, trained using the original Pix2Pix architecture, was selected at the indicated epoch.
      \item{e} Refiner1 and Refiner2, trained using the modified Pix2Pix model with a secondary structure discriminator, were selected at the listed epochs. For Ala\textsubscript{10}, only the original Pix2Pix refiner was used.
  \end{tablenotes}
  
\newpage
\begin{sidewaystable}[p]  

\centering
\renewcommand{\tablename}{Table S\hskip-\the\fontdimen2\font\space }
\caption{Backbone‑geometry descriptors commonly used in protein validation.}
\label{tab:s2}
\resizebox{\textwidth}{!}{%
\begin{tabular}{|l|p{5.2cm}|p{7.8cm}|}
\hline
\textbf{Geometric feature} & \textbf{Definition} & \textbf{Structural role / diagnostic use} \\ \hline
C--C$_\alpha$ bond   & Carbonyl~C to C$_\alpha$                                                          & Part of the peptide backbone; large deviations indicate refinement errors. \\ \hline
N--C bond            & Peptide~N to carbonyl~C                                      & Planar and partially double‑bonded; large deviations indicate geometry errors. \\ \hline
C$_\alpha$--N bond   & C$_\alpha$ to peptide~N                                                          & Completes the backbone triad of bond lengths; large deviations indicate stereochemical problems. \\ \hline
N--C$_\alpha$--C angle & Backbone bond angle at C$_\alpha$                                              & Governs local chain direction; large deviations indicate strained conformations. \\ \hline
$\phi$ torsion       & C$_{i-1}$--N$_i$--C$_{\alpha i}$--C$_i$                                          & One of the two principal dihedrals defining backbone conformation; plotted on a Ramachandran map. large deviations indicate wrong geometry errors in secondary structure \\ \hline
$\psi$ torsion       & N$_i$--C$_{\alpha i}$--C$_i$--N$_{i+1}$                                          & Partner to $\phi$; Similar to $\phi$. \\ \hline
\end{tabular}}
\end{sidewaystable}
\clearpage

\newpage
\begin{sidewaystable}[htbp]
  \centering
  \renewcommand{\tablename}{Table S\hskip-\the\fontdimen2\font\space}
    \caption{Whole‑Backbone Performance Evaluation of Rigid Protein Structures}
    \label{tab:whole}
    \resizebox{\linewidth}{!}{%
        \begin{tabular}{cl|ll|llllll|}
        \cline{3-10}
        \multicolumn{1}{l}{}                                                          &                                                   & \multicolumn{2}{c|}{\cellcolor[HTML]{EFEFEF}\textbf{Benchmark\textsuperscript{a}}}                                             & \multicolumn{6}{c|}{\cellcolor[HTML]{C0C0C0}MoDyGAN\textsuperscript{b}}                                                                                                                                                                                                                                                                                                                                                                                                                        \\ \hline
        \rowcolor[HTML]{C0C0C0} 
        \multicolumn{1}{|l|}{\cellcolor[HTML]{FFFFFF}\textbf{Protein}}                & \cellcolor[HTML]{FFFFFF}\textbf{Metrics}          & \multicolumn{1}{l|}{\cellcolor[HTML]{EFEFEF}\textbf{Simulation}} & \cellcolor[HTML]{EFEFEF}\textbf{Decoder} & \multicolumn{1}{l|}{\cellcolor[HTML]{C0C0C0}\textbf{Simulation}}                  & \multicolumn{1}{l|}{\cellcolor[HTML]{C0C0C0}\textbf{Recover}}                     & \multicolumn{1}{l|}{\cellcolor[HTML]{C0C0C0}\textbf{MoDyGAN (Orig.)}}               & \multicolumn{1}{l|}{\cellcolor[HTML]{C0C0C0}\textbf{MoDyGAN (Refiner1)}}            & \multicolumn{1}{l|}{\cellcolor[HTML]{C0C0C0}\textbf{MoDyGAN (Refiner2)}}            & \textbf{MoDyGAN ((Ensemble)}            \\
\hline
      \end{tabular}%
    }
    \begin{tablenotes}[flushleft]
      \footnotesize         
          \item[] Evaluation on RMSD, bond and angle analysis and backbone energy are reported as mean ± std. RF accuracy is evaluated on the \textit{validation dataset}. RF accept rates are evaluated across all generated conformations. Also see Supplementary Table S \ref{tab:1poa_part}- S \ref{tab:1bmr_part}.
         
          \item[a] a:The benchmark includes MD simulation samples from [6] and [6]-generated conformations. MD simulation samples serve as the baseline  for comparing [6]-generated conformations. Metrics for 1BMR are omitted due to no pre‑trained autoencoder. RF accuracy fields are blank because no pre‑trained RF classifiers from [6].
       
          \item[b] b:It includes all variations. MD simulation samples are obtained from the process described in Supplementary Test B. It serves as the baseline for comparing MoDyGAN-generated conformations.

    \end{tablenotes}

\end{sidewaystable}
\clearpage
\newpage
\begin{sidewaystable}[htbp]
  \centering
    \renewcommand{\tablename}{Table S\hskip-\the\fontdimen2\font\space}
    \caption{Secondary Segment Performance Evaluation of 1POA}
    \label{tab:1poa_part}
    \resizebox{\linewidth}{!}{%

\begin{tabular}{cl|ll|llllll}
        \cline{3-4}
        \multicolumn{1}{l}{}                                                            &                                                   & \multicolumn{2}{c|}{\cellcolor[HTML]{EFEFEF}\textbf{Benchmark}}                                             & \multicolumn{6}{c}{\cellcolor[HTML]{C0C0C0}\textbf{MoDyGAN}}                                                                                                                                                                                                                                                                                                                                                                                                                                                \\ \hline
        \rowcolor[HTML]{C0C0C0} 
        \multicolumn{1}{|l|}{\cellcolor[HTML]{FFFFFF}\textbf{Segments}}                 & \cellcolor[HTML]{FFFFFF}\textbf{Metrics}          & \multicolumn{1}{l|}{\cellcolor[HTML]{EFEFEF}\textbf{Simulation}} & \cellcolor[HTML]{EFEFEF}\textbf{Decoder} & \multicolumn{1}{l|}{\cellcolor[HTML]{C0C0C0}\textbf{Simulation}}                & \multicolumn{1}{l|}{\cellcolor[HTML]{C0C0C0}\textbf{Recover}}                   & \multicolumn{1}{l|}{\cellcolor[HTML]{C0C0C0}\textbf{MoDyGAN (Orig.)}}             & \multicolumn{1}{l|}{\cellcolor[HTML]{C0C0C0}\textbf{MoDyGAN (Refiner1)}}          & \multicolumn{1}{l|}{\cellcolor[HTML]{C0C0C0}\textbf{MoDyGAN (Refiner2)}}          & \multicolumn{1}{l|}{\cellcolor[HTML]{C0C0C0}\textbf{MoDyGAN (Ensembling)}}        \\
\hline
        \end{tabular}
    }
    \begin{tablenotes}[flushleft]
      \footnotesize
        \item[] Related to Supplementary Table S \ref{tab:whole}. 1POA is divided into four secondary structure regions, and all metrics are computed specifically for these regions to assess local structural fidelity.
    \end{tablenotes}

\end{sidewaystable}
\clearpage

\newpage
\begin{sidewaystable}[htbp]
  \centering
    \renewcommand{\tablename}{Table S\hskip-\the\fontdimen2\font\space}
    \caption{Secondary Segment Performance Evaluation of 2WJ7 (part~1~of~2).}
    \label{tab:2wj7_part1}
    \resizebox{\linewidth}{!}{%

        \begin{tabular}{cl|ll|llllll|}
        \cline{3-10}
        \multicolumn{1}{l}{}                                                            &                                                   & \multicolumn{2}{c|}{\cellcolor[HTML]{EFEFEF}\textbf{Benchmark}}                                             & \multicolumn{6}{c|}{\cellcolor[HTML]{C0C0C0}\textbf{MoDyGAN}}                                                                                                                                                                                                                                                                                                                                                                                                                \\ \hline
        \rowcolor[HTML]{C0C0C0} 
        \multicolumn{1}{|l|}{\cellcolor[HTML]{FFFFFF}\textbf{Segments}}                 & \cellcolor[HTML]{FFFFFF}\textbf{Metrics}          & \multicolumn{1}{l|}{\cellcolor[HTML]{EFEFEF}\textbf{Simulation}} & \cellcolor[HTML]{EFEFEF}\textbf{Decoder} & \multicolumn{1}{l|}{\cellcolor[HTML]{C0C0C0}\textbf{Simulation}}                  & \multicolumn{1}{l|}{\cellcolor[HTML]{C0C0C0}\textbf{Recover}}                     & \multicolumn{1}{l|}{\cellcolor[HTML]{C0C0C0}\textbf{MoDyGAN (Orig.)}}               & \multicolumn{1}{l|}{\cellcolor[HTML]{C0C0C0}\textbf{MoDyGAN (Refiner1)}}            & \multicolumn{1}{l|}{\cellcolor[HTML]{C0C0C0}\textbf{MoDyGAN (Refiner2)}}            & \textbf{MoDyGAN (Ensembling)}            \\
\hline
        \end{tabular}
    }
    \begin{tablenotes}[flushleft]
      \footnotesize
        \item[] Related to Supplementary Table S \ref{tab:whole}. 2WJ7 is divided into eight secondary structure regions. Metrics for segments 1–4 to assess local structural fidelity.
    \end{tablenotes}

\end{sidewaystable}
\clearpage

\newpage
\begin{sidewaystable}[htbp]
  \centering
    \renewcommand{\tablename}{Table S\hskip-\the\fontdimen2\font\space}
    \caption{Secondary Segment Performance Evaluation of 2WJ7 (part~2~of~2).}
    \label{tab:2wj7_part2}
    \resizebox{\linewidth}{!}{%
        \begin{tabular}{cl|ll|llllll|}
\cline{3-10}
\multicolumn{1}{l}{}                                                           &                                                   & \multicolumn{2}{c|}{\cellcolor[HTML]{EFEFEF}\textbf{Benchmark}}                                             & \multicolumn{6}{c|}{\cellcolor[HTML]{C0C0C0}\textbf{MoDyGAN}}                                                                                                                                                                                                                                                                                                                                                                                                                 \\ \hline
\rowcolor[HTML]{C0C0C0} 
\multicolumn{1}{|l|}{\cellcolor[HTML]{FFFFFF}\textbf{Segments}}                & \cellcolor[HTML]{FFFFFF}\textbf{Metrics}          & \multicolumn{1}{l|}{\cellcolor[HTML]{EFEFEF}\textbf{Simulation}} & \cellcolor[HTML]{EFEFEF}\textbf{Decoder} & \multicolumn{1}{l|}{\cellcolor[HTML]{C0C0C0}\textbf{Simulation}}                  & \multicolumn{1}{l|}{\cellcolor[HTML]{C0C0C0}\textbf{Recover}}                     & \multicolumn{1}{l|}{\cellcolor[HTML]{C0C0C0}\textbf{MoDyGAN (Orig.)}}               & \multicolumn{1}{l|}{\cellcolor[HTML]{C0C0C0}\textbf{MoDyGAN (Refiner1)}}            & \multicolumn{1}{l|}{\cellcolor[HTML]{C0C0C0}\textbf{MoDyGAN (Refiner2)}}            & \textbf{MoDyGAN (Ensembling)}             \\
\hline
\end{tabular}
  }
    \begin{tablenotes}[flushleft]
      \footnotesize
        \item[] Related to Supplementary Table S \ref{tab:whole}. Supplementary Table S \ref{tab:2wj7_part1} cont. Metrics for segments 5-8 to assess local structural fidelity.
    \end{tablenotes}

\end{sidewaystable}
\clearpage

\newpage
\begin{sidewaystable}[htbp]
  \centering
    \renewcommand{\tablename}{Table S\hskip-\the\fontdimen2\font\space}
    \captionsetup{justification=centering,singlelinecheck=false}
    \caption{Secondary Segment Performance Evaluation of 1BMR}
    \label{tab:1bmr_part}
    \resizebox{\linewidth}{!}{%

        \begin{tabular}{|c|l|l|l|l|l|l|l|}
        \hline
        \rowcolor[HTML]{C0C0C0} 
        \multicolumn{1}{|l|}{\cellcolor[HTML]{FFFFFF}\textbf{Segments}} & \cellcolor[HTML]{FFFFFF}\textbf{Metrics}          & \textbf{Simulation}                  & \textbf{Recover}                     & \textbf{MoDyGAN (Orig.)}               & \textbf{MoDyGAN (Refiner1)}            & \textbf{MoDyGAN (Refiner2)}            & \textbf{MoDyGAN (Ensembling)}             \\ \hline
        \rowcolor[HTML]{9B9B9B} 
        \cellcolor[HTML]{FFFFFF}                                        & \cellcolor[HTML]{FFFFFF}C-CA (\AA)                & 1.51 ± 0.04                          & 1.13 ± 0.24                          & 1.46 ± 0.08                          & 1.42 ± 0.10                          & 1.45 ± 0.08                          & \cellcolor[HTML]{9B9B9B}1.45 ± 0.08     \\ \cline{2-8} 
        \rowcolor[HTML]{9B9B9B} 
        \cellcolor[HTML]{FFFFFF}                                        & \cellcolor[HTML]{FFFFFF}N-C (\AA)                 & 1.35 ± 0.03                          & 1.02 ± 0.19                          & 1.30 ± 0.06                          & 1.30 ± 0.08                          & 1.31 ± 0.06                          & \cellcolor[HTML]{9B9B9B}1.31 ± 0.06     \\ \cline{2-8} 
        \rowcolor[HTML]{9B9B9B} 
        \cellcolor[HTML]{FFFFFF}                                        & \cellcolor[HTML]{FFFFFF}CA-N (\AA)                & 1.46 ± 0.04                          & 1.22 ± 0.26                          & 1.45 ± 0.12                          & 1.38 ± 0.09                          & 1.41 ± 0.08                          & \cellcolor[HTML]{9B9B9B}1.40 ± 0.08     \\ \cline{2-8} 
        \rowcolor[HTML]{9B9B9B} 
        \cellcolor[HTML]{FFFFFF}                                        & \cellcolor[HTML]{FFFFFF}N-CA-C (\si{\degree})              & 110.98 ± 4.35                        & 139.92 ± 20.05                       & 111.14 ± 6.01                        & 116.72 ± 6.63                        & 115.60 ± 6.95                        & \cellcolor[HTML]{9B9B9B}115.47 ± 7.02   \\ \cline{2-8} 
        \rowcolor[HTML]{9B9B9B} 
        \cellcolor[HTML]{FFFFFF}                                        & \cellcolor[HTML]{FFFFFF}Phi (\si{\degree})        & -54.37 ± 132.39                      & -45.21 ± 124.34                      & -30.14 ± 138.77                      & -35.02 ± 140.03                      & -34.72 ± 141.21                      & \cellcolor[HTML]{9B9B9B}-34.65 ± 141.03 \\ \cline{2-8} 
        \rowcolor[HTML]{9B9B9B} 
        \cellcolor[HTML]{FFFFFF}                                        & \cellcolor[HTML]{FFFFFF}Psi (\si{\degree})        & 146.74 ± 35.14                       & 74.20 ± 113.30                       & 147.79 ± 11.85                       & 148.96 ± 17.14                       & 147.61 ± 14.13                       & \cellcolor[HTML]{9B9B9B}147.72 ± 14.32  \\ \cline{2-8} 
        \rowcolor[HTML]{9B9B9B} 
        \cellcolor[HTML]{FFFFFF}                                        & \cellcolor[HTML]{FFFFFF}RMSD (\AA)                & 0.33 ± 0.06                        & 0.56 ± 0.23                        & 0.30 ± 0.12                        & 0.31 ± 0.13                        & 0.30 ± 0.12                        & 0.30 ± 0.12                           \\ \cline{2-8} 
        \rowcolor[HTML]{9B9B9B} 
        \multirow{-8}{*}{\cellcolor[HTML]{FFFFFF}\textbf{BETA1}}        & \cellcolor[HTML]{FFFFFF}RF Accuracy / Accept (\%) & 100.00                               & 3.09                                 & 78.57                                & 86.16                                & 91.93                                & 90.70                                   \\ \hline
        \rowcolor[HTML]{C0C0C0} 
        \cellcolor[HTML]{FFFFFF}                                        & \cellcolor[HTML]{FFFFFF}C-CA (\AA)                & 1.52 ± 0.04                          & 1.18 ± 0.24                          & 1.47 ± 0.08                          & 1.46 ± 0.09                          & 1.47 ± 0.08                          & 1.47 ± 0.08                             \\ \cline{2-8} 
        \rowcolor[HTML]{C0C0C0} 
        \cellcolor[HTML]{FFFFFF}                                        & \cellcolor[HTML]{FFFFFF}N-C (\AA)                 & 1.35 ± 0.03                          & 1.11 ± 0.22                          & 1.30 ± 0.06                          & 1.29 ± 0.09                          & 1.32 ± 0.07                          & 1.32 ± 0.07                             \\ \cline{2-8} 
        \rowcolor[HTML]{C0C0C0} 
        \cellcolor[HTML]{FFFFFF}                                        & \cellcolor[HTML]{FFFFFF}CA-N (\AA)                & 1.46 ± 0.03                          & 1.21 ± 0.23                          & 1.44 ± 0.09                          & 1.40 ± 0.09                          & 1.42 ± 0.08                          & 1.41 ± 0.08                             \\ \cline{2-8} 
        \rowcolor[HTML]{C0C0C0} 
        \cellcolor[HTML]{FFFFFF}                                        & \cellcolor[HTML]{FFFFFF}N-CA-C (\si{\degree})              & 111.65 ± 4.33                        & 136.52 ± 18.39                       & 113.84 ± 5.98                        & 115.75 ± 5.84                        & 115.86 ± 6.11                        & 115.68 ± 6.14                           \\ \cline{2-8} 
        \rowcolor[HTML]{C0C0C0} 
        \cellcolor[HTML]{FFFFFF}                                        & \cellcolor[HTML]{FFFFFF}Phi (\si{\degree})        & -73.82 ± 85.67                       & -59.99 ± 87.75                       & -65.72 ± 91.78                       & -67.03 ± 91.45                       & -65.43 ± 91.91                       & -65.55 ± 91.84                          \\ \cline{2-8} 
        \rowcolor[HTML]{C0C0C0} 
        \cellcolor[HTML]{FFFFFF}                                        & \cellcolor[HTML]{FFFFFF}Psi (\si{\degree})        & 121.51 ± 87.08                       & 78.96 ± 109.94                       & 109.01 ± 103.01                      & 108.93 ± 102.72                      & 106.75 ± 103.30                      & 106.86 ± 103.37                         \\ \cline{2-8} 
        \rowcolor[HTML]{C0C0C0} 
        \cellcolor[HTML]{FFFFFF}                                        & \cellcolor[HTML]{FFFFFF}RMSD (\AA)                & 0.35 ± 0.06                        & 0.80 ± 0.35                        & 0.40 ± 0.18                        & 0.43 ± 0.18                        & 0.42 ± 0.17                        & 0.41 ± 0.17                           \\ \cline{2-8} 
        \rowcolor[HTML]{C0C0C0} 
        \multirow{-8}{*}{\cellcolor[HTML]{FFFFFF}\textbf{ALPHA1}}       & \cellcolor[HTML]{FFFFFF}RF Accuracy / Accept (\%) & 100.00                               & 0.51                                 & 75.35                                & 47.92                                & 47.85                                & 49.65                                   \\ \hline
        \rowcolor[HTML]{9B9B9B} 
        \cellcolor[HTML]{FFFFFF}                                        & \cellcolor[HTML]{FFFFFF}C-CA (\AA)                & 1.52 ± 0.04                          & 1.17 ± 0.24                          & 1.47 ± 0.08                          & 1.44 ± 0.10                          & 1.46 ± 0.08                          & \cellcolor[HTML]{9B9B9B}1.46 ± 0.08     \\ \cline{2-8} 
        \rowcolor[HTML]{9B9B9B} 
        \cellcolor[HTML]{FFFFFF}                                        & \cellcolor[HTML]{FFFFFF}N-C (\AA)                 & 1.35 ± 0.03                          & 1.09 ± 0.22                          & 1.29 ± 0.06                          & 1.28 ± 0.08                          & 1.30 ± 0.07                          & \cellcolor[HTML]{9B9B9B}1.30 ± 0.07     \\ \cline{2-8} 
        \rowcolor[HTML]{9B9B9B} 
        \cellcolor[HTML]{FFFFFF}                                        & \cellcolor[HTML]{FFFFFF}CA-N (\AA)                & 1.46 ± 0.04                          & 1.19 ± 0.24                          & 1.45 ± 0.11                          & 1.40 ± 0.09                          & 1.42 ± 0.08                          & \cellcolor[HTML]{9B9B9B}1.41 ± 0.08     \\ \cline{2-8} 
        \rowcolor[HTML]{9B9B9B} 
        \cellcolor[HTML]{FFFFFF}                                        & \cellcolor[HTML]{FFFFFF}N-CA-C (\si{\degree})              & 111.29 ± 4.38                        & 137.75 ± 17.76                       & 112.60 ± 6.43                        & 115.69 ± 6.35                        & 114.76 ± 6.68                        & \cellcolor[HTML]{9B9B9B}114.63 ± 6.73   \\ \cline{2-8} 
        \rowcolor[HTML]{9B9B9B} 
        \cellcolor[HTML]{FFFFFF}                                        & \cellcolor[HTML]{FFFFFF}Phi (\si{\degree})        & -63.07 ± 103.52                      & -47.56 ± 98.91                       & -48.98 ± 110.04                      & -51.72 ± 110.48                      & -52.27 ± 111.52                      & \cellcolor[HTML]{9B9B9B}-52.21 ± 111.41 \\ \cline{2-8} 
        \rowcolor[HTML]{9B9B9B} 
        \cellcolor[HTML]{FFFFFF}                                        & \cellcolor[HTML]{FFFFFF}Psi (\si{\degree})        & 105.70 ± 104.55                      & 58.74 ± 120.34                       & 84.34 ± 123.58                       & 85.04 ± 123.47                       & 82.27 ± 124.07                       & \cellcolor[HTML]{9B9B9B}82.43 ± 124.13  \\ \cline{2-8} 
        \rowcolor[HTML]{9B9B9B} 
        \cellcolor[HTML]{FFFFFF}                                        & \cellcolor[HTML]{FFFFFF}RMSD (\AA)                & {\color[HTML]{000000} 0.43 ± 0.12} & {\color[HTML]{000000} 0.80 ± 0.41} & {\color[HTML]{000000} 0.45 ± 0.21} & {\color[HTML]{000000} 0.46 ± 0.21} & {\color[HTML]{000000} 0.45 ± 0.22} & {\color[HTML]{000000} 0.45 ± 0.21}    \\ \cline{2-8} 
        \rowcolor[HTML]{9B9B9B} 
        \multirow{-8}{*}{\cellcolor[HTML]{FFFFFF}\textbf{BETA2}}        & \cellcolor[HTML]{FFFFFF}RF Accuracy / Accept (\%) & 100.00                               & 2.44                                 & 70.61                                & 53.10                                & 52.77                                & 80.58                                   \\ \hline
        \rowcolor[HTML]{C0C0C0} 
        \cellcolor[HTML]{FFFFFF}                                        & \cellcolor[HTML]{FFFFFF}C-CA (\AA)                & 1.52 ± 0.04                          & 1.20 ± 0.24                          & 1.47 ± 0.07                          & 1.45 ± 0.09                          & 1.47 ± 0.08                          & 1.47 ± 0.08                             \\ \cline{2-8} 
        \rowcolor[HTML]{C0C0C0} 
        \cellcolor[HTML]{FFFFFF}                                        & \cellcolor[HTML]{FFFFFF}N-C (\AA)                 & 1.35 ± 0.03                          & 1.12 ± 0.22                          & 1.30 ± 0.06                          & 1.29 ± 0.08                          & 1.32 ± 0.07                          & 1.31 ± 0.07                             \\ \cline{2-8} 
        \rowcolor[HTML]{C0C0C0} 
        \cellcolor[HTML]{FFFFFF}                                        & \cellcolor[HTML]{FFFFFF}CA-N (\AA)                & 1.46 ± 0.04                          & 1.20 ± 0.24                          & 1.44 ± 0.10                          & 1.40 ± 0.09                          & 1.42 ± 0.08                          & 1.41 ± 0.08                             \\ \cline{2-8} 
        \rowcolor[HTML]{C0C0C0} 
        \cellcolor[HTML]{FFFFFF}                                        & \cellcolor[HTML]{FFFFFF}N-CA-C (\si{\degree})              & 111.22 ± 4.20                        & 138.45 ± 16.68                       & 113.60 ± 6.17                        & 115.50 ± 5.96                        & 115.41 ± 6.21                        & 115.19 ± 6.22                           \\ \cline{2-8} 
        \rowcolor[HTML]{C0C0C0} 
        \cellcolor[HTML]{FFFFFF}                                        & \cellcolor[HTML]{FFFFFF}Phi (\si{\degree})        & -75.67 ± 91.27                       & -61.94 ± 91.17                       & -66.30 ± 98.10                       & -66.89 ± 97.70                       & -66.30 ± 98.21                       & -66.36 ± 98.10                          \\ \cline{2-8} 
        \rowcolor[HTML]{C0C0C0} 
        \cellcolor[HTML]{FFFFFF}                                        & \cellcolor[HTML]{FFFFFF}Psi (\si{\degree})        & 117.01 ± 92.12                       & 75.14 ± 114.94                       & 102.61 ± 108.56                      & 102.50 ± 108.21                      & 100.34 ± 108.90                      & 100.51 ± 108.90                         \\ \cline{2-8} 
        \rowcolor[HTML]{C0C0C0} 
        \cellcolor[HTML]{FFFFFF}                                        & \cellcolor[HTML]{FFFFFF}RMSD (\AA)                & 0.55 ± 0.13                        & 0.97 ± 0.46                        & 0.60 ± 0.23                        & 0.61 ± 0.24                        & 0.61 ± 0.22                        & 0.61 ± 0.23                           \\ \cline{2-8} 
        \rowcolor[HTML]{C0C0C0} 
        \multirow{-8}{*}{\cellcolor[HTML]{FFFFFF}\textbf{BETA3}}        & \cellcolor[HTML]{FFFFFF}RF Accuracy / Accept (\%) & 100.00                               & 0.01                                 & 74.78                                & 58.89                                & 76.73                                & 85.36                                   \\ \hline
        \end{tabular}
  }
    \begin{tablenotes}[flushleft]
      \footnotesize
        \item[] Related to Supplementary Table S \ref{tab:whole}. 1BMR is divided into four secondary structure regions, and all metrics are computed specifically for these regions to assess local structural fidelity.
    \end{tablenotes}

\end{sidewaystable}
\clearpage

\newpage
\begin{sidewaystable}[htbp]
  \centering
  \renewcommand{\tablename}{Table S\hskip-\the\fontdimen2\font\space}
    \caption{Whole‑Backbone Performance Evaluation of Ala\textsubscript{10}}
    \label{tab:10ala_backbone_comparison}
    \resizebox{\linewidth}{!}{%
\begin{tabular}{|l|l|
>{\columncolor[HTML]{EFEFEF}}l 
>{\columncolor[HTML]{EFEFEF}}l |
>{\columncolor[HTML]{C0C0C0}}l 
>{\columncolor[HTML]{C0C0C0}}l 
>{\columncolor[HTML]{C0C0C0}}l 
>{\columncolor[HTML]{C0C0C0}}l |
>{\columncolor[HTML]{9B9B9B}}l 
>{\columncolor[HTML]{9B9B9B}}l 
>{\columncolor[HTML]{9B9B9B}}l 
>{\columncolor[HTML]{9B9B9B}}l |}
\hline
\multicolumn{1}{|c|}{}                         & \multicolumn{1}{c|}{}                                                                              & \multicolumn{2}{c|}{\cellcolor[HTML]{EFEFEF}RMSD Only\textsuperscript{b}}                                                              & \multicolumn{4}{c|}{\cellcolor[HTML]{C0C0C0}RMSD + Angle Valid\textsuperscript{c}}                                                                                                                                                                                     & \multicolumn{4}{c|}{\cellcolor[HTML]{9B9B9B}RMSD + Angle + Energy Valid \textsuperscript{d}}                                                                                                                                                                          \\ \cline{3-12} 
\multicolumn{1}{|c|}{\multirow{-2}{*}{States}} & \multicolumn{1}{c|}{\multirow{-2}{*}{\begin{tabular}[c]{@{}c@{}}Baseline E\textsuperscript{a} \\ (REU)\end{tabular}}} & \multicolumn{1}{c|}{\cellcolor[HTML]{EFEFEF}RMSD (\AA)}       & \multicolumn{1}{c|}{\cellcolor[HTML]{EFEFEF}Amount} & \multicolumn{1}{c|}{\cellcolor[HTML]{C0C0C0}RMSD (\AA)}      & \multicolumn{1}{c|}{\cellcolor[HTML]{C0C0C0}EMD (IS)}        & \multicolumn{1}{c|}{\cellcolor[HTML]{C0C0C0}Energy (REU)}       & \multicolumn{1}{c|}{\cellcolor[HTML]{C0C0C0}Amount} & \multicolumn{1}{c|}{\cellcolor[HTML]{9B9B9B}RMSD (\AA)}      & \multicolumn{1}{c|}{\cellcolor[HTML]{9B9B9B}EMD (IS)}        & \multicolumn{1}{c|}{\cellcolor[HTML]{9B9B9B}Energy (REU)}     & \multicolumn{1}{c|}{\cellcolor[HTML]{9B9B9B}Amount} \\ \hline
State0                                         & 38.228                                                                                             & \multicolumn{1}{l|}{\cellcolor[HTML]{EFEFEF}3.131 ± 0.303}  & 3317                                                & \multicolumn{1}{l|}{\cellcolor[HTML]{C0C0C0}3.012 ± 0.449} & \multicolumn{1}{l|}{\cellcolor[HTML]{C0C0C0}0.030 ± 0.010} & \multicolumn{1}{l|}{\cellcolor[HTML]{C0C0C0}140.387 ± 39.899} & 528                                                 & \multicolumn{1}{l|}{\cellcolor[HTML]{9B9B9B}3.104 ± 0.084} & \multicolumn{1}{l|}{\cellcolor[HTML]{9B9B9B}0.029 ± 0.006} & \multicolumn{1}{l|}{\cellcolor[HTML]{9B9B9B}29.158 ± 5.099} & 3                                                   \\ \hline
State1                                         & 20.971                                                                                             & \multicolumn{1}{l|}{\cellcolor[HTML]{EFEFEF}3.411 ± 0.323,} & 698                                                 & \multicolumn{1}{l|}{\cellcolor[HTML]{C0C0C0}3.238 ± 0.421} & \multicolumn{1}{l|}{\cellcolor[HTML]{C0C0C0}0.043 ± 0.011} & \multicolumn{1}{l|}{\cellcolor[HTML]{C0C0C0}146.403 ± 50.161} & 66                                                  & \multicolumn{1}{l|}{\cellcolor[HTML]{9B9B9B}1.712 ± 0.000} & \multicolumn{1}{l|}{\cellcolor[HTML]{9B9B9B}0.049 ± 0.000} & \multicolumn{1}{l|}{\cellcolor[HTML]{9B9B9B}9.132 ± 0.000}  & 1                                                   \\ \hline
State2                                         & 15.013                                                                                             & \multicolumn{1}{l|}{\cellcolor[HTML]{EFEFEF}2.061 ± 0.857}  & 52                                                  & \multicolumn{1}{l|}{\cellcolor[HTML]{C0C0C0}1.786 ± 0.771} & \multicolumn{1}{l|}{\cellcolor[HTML]{C0C0C0}0.042 ± 0.015} & \multicolumn{1}{l|}{\cellcolor[HTML]{C0C0C0}105.352 ± 67.341} & 26                                                  & \multicolumn{1}{l|}{\cellcolor[HTML]{9B9B9B}/}               & \multicolumn{1}{l|}{\cellcolor[HTML]{9B9B9B}/}               & \multicolumn{1}{l|}{\cellcolor[HTML]{9B9B9B}/}                & 0                                                   \\ \hline
State3                                         & 33.813                                                                                             & \multicolumn{1}{l|}{\cellcolor[HTML]{EFEFEF}2.846 ± 0.756}  & 72                                                  & \multicolumn{1}{l|}{\cellcolor[HTML]{C0C0C0}2.221 ± 0.825} & \multicolumn{1}{l|}{\cellcolor[HTML]{C0C0C0}0.047 ± 0.017} & \multicolumn{1}{l|}{\cellcolor[HTML]{C0C0C0}120.120 ± 65.603} & 12                                                  & \multicolumn{1}{l|}{\cellcolor[HTML]{9B9B9B}1.613 ± 0.000} & \multicolumn{1}{l|}{\cellcolor[HTML]{9B9B9B}0.049 ± 0.000} & \multicolumn{1}{l|}{\cellcolor[HTML]{9B9B9B}23.619 ± 0.000} & 1                                                   \\ \hline
State4                                         & 11.346                                                                                             & \multicolumn{1}{l|}{\cellcolor[HTML]{EFEFEF}2.021 ± 0.715}  & 69                                                  & \multicolumn{1}{l|}{\cellcolor[HTML]{C0C0C0}1.789 ± 0.624} & \multicolumn{1}{l|}{\cellcolor[HTML]{C0C0C0}0.037 ± 0.019} & \multicolumn{1}{l|}{\cellcolor[HTML]{C0C0C0}112.736 ± 59.121} & 19                                                  & \multicolumn{1}{l|}{\cellcolor[HTML]{9B9B9B}/}               & \multicolumn{1}{l|}{\cellcolor[HTML]{9B9B9B}/}               & \multicolumn{1}{l|}{\cellcolor[HTML]{9B9B9B}/}                & 0                                                   \\ \hline
State5                                         & 16.625                                                                                             & \multicolumn{1}{l|}{\cellcolor[HTML]{EFEFEF}1.172 ± 0.640}  & 1110                                                & \multicolumn{1}{l|}{\cellcolor[HTML]{C0C0C0}1.295 ± 0.669} & \multicolumn{1}{l|}{\cellcolor[HTML]{C0C0C0}0.041 ± 0.018} & \multicolumn{1}{l|}{\cellcolor[HTML]{C0C0C0}70.033 ± 67.535}  & 283                                                 & \multicolumn{1}{l|}{\cellcolor[HTML]{9B9B9B}0.798 ± 0.123} & \multicolumn{1}{l|}{\cellcolor[HTML]{9B9B9B}0.049 ± 0.019} & \multicolumn{1}{l|}{\cellcolor[HTML]{9B9B9B}10.499 ± 3.956} & 73                                                  \\ \hline
State6                                         & 17.868                                                                                             & \multicolumn{1}{l|}{\cellcolor[HTML]{EFEFEF}0.882 ± 0.316}  & 4001                                                & \multicolumn{1}{l|}{\cellcolor[HTML]{C0C0C0}1.010 ± 0.446} & \multicolumn{1}{l|}{\cellcolor[HTML]{C0C0C0}0.039 ± 0.017} & \multicolumn{1}{l|}{\cellcolor[HTML]{C0C0C0}46.235 ± 50.682}  & 913                                                 & \multicolumn{1}{l|}{\cellcolor[HTML]{9B9B9B}0.789 ± 0.110} & \multicolumn{1}{l|}{\cellcolor[HTML]{9B9B9B}0.042 ± 0.018} & \multicolumn{1}{l|}{\cellcolor[HTML]{9B9B9B}0.607 ± 4.179}  & 333                                                 \\ \hline
State7                                         & 4.720                                                                                              & \multicolumn{1}{l|}{\cellcolor[HTML]{EFEFEF}1.364 ± 0.293}  & 5                                                   & \multicolumn{1}{l|}{\cellcolor[HTML]{C0C0C0}1.572 ± 0.000} & \multicolumn{1}{l|}{\cellcolor[HTML]{C0C0C0}0.062 ± 0.000} & \multicolumn{1}{l|}{\cellcolor[HTML]{C0C0C0}111.888 ± 0.000}  & 1                                                   & \multicolumn{1}{l|}{\cellcolor[HTML]{9B9B9B}/}               & \multicolumn{1}{l|}{\cellcolor[HTML]{9B9B9B}/}               & \multicolumn{1}{l|}{\cellcolor[HTML]{9B9B9B}/}                & 0                                                   \\ \hline
State8                                         & 19.420                                                                                             & \multicolumn{1}{l|}{\cellcolor[HTML]{EFEFEF}2.546 ± 0.645}  & 154                                                 & \multicolumn{1}{l|}{\cellcolor[HTML]{C0C0C0}1.744 ± 0.367} & \multicolumn{1}{l|}{\cellcolor[HTML]{C0C0C0}0.039 ± 0.018} & \multicolumn{1}{l|}{\cellcolor[HTML]{C0C0C0}115.297 ± 55.052} & 31                                                  & \multicolumn{1}{l|}{\cellcolor[HTML]{9B9B9B}1.546 ± 0.000} & \multicolumn{1}{l|}{\cellcolor[HTML]{9B9B9B}0.062 ± 0.000} & \multicolumn{1}{l|}{\cellcolor[HTML]{9B9B9B}19.161 ± 0.000} & 1                                                   \\ \hline
State9                                         & 18.040                                                                                             & \multicolumn{1}{l|}{\cellcolor[HTML]{EFEFEF}1.395 ± 0.418}  & 19                                                  & \multicolumn{1}{l|}{\cellcolor[HTML]{C0C0C0}1.470 ± 0.188} & \multicolumn{1}{l|}{\cellcolor[HTML]{C0C0C0}0.049 ± 0.000} & \multicolumn{1}{l|}{\cellcolor[HTML]{C0C0C0}146.393 ± 63.715} & 3                                                   & \multicolumn{1}{l|}{\cellcolor[HTML]{9B9B9B}/}               & \multicolumn{1}{l|}{\cellcolor[HTML]{9B9B9B}/}               & \multicolumn{1}{l|}{\cellcolor[HTML]{9B9B9B}/}                & 0                                                   \\ \hline
State10                                        & 8.178                                                                                              & \multicolumn{1}{l|}{\cellcolor[HTML]{EFEFEF}1.995 ± 0.453}  & 18                                                  & \multicolumn{1}{l|}{\cellcolor[HTML]{C0C0C0}1.897 ± 0.187} & \multicolumn{1}{l|}{\cellcolor[HTML]{C0C0C0}0.043 ± 0.006} & \multicolumn{1}{l|}{\cellcolor[HTML]{C0C0C0}154.067 ± 18.469} & 2                                                   & \multicolumn{1}{l|}{\cellcolor[HTML]{9B9B9B}/}               & \multicolumn{1}{l|}{\cellcolor[HTML]{9B9B9B}/}               & \multicolumn{1}{l|}{\cellcolor[HTML]{9B9B9B}/}                & 0                                                   \\ \hline
State11                                        & 15.374                                                                                             & \multicolumn{1}{l|}{\cellcolor[HTML]{EFEFEF}1.505 ± 0.431}  & 2                                                   & \multicolumn{1}{l|}{\cellcolor[HTML]{C0C0C0}/}               & \multicolumn{1}{l|}{\cellcolor[HTML]{C0C0C0}/}               & \multicolumn{1}{l|}{\cellcolor[HTML]{C0C0C0}/}                  & 0                                                   & \multicolumn{1}{l|}{\cellcolor[HTML]{9B9B9B}/}               & \multicolumn{1}{l|}{\cellcolor[HTML]{9B9B9B}/}               & \multicolumn{1}{l|}{\cellcolor[HTML]{9B9B9B}/}                & 0                                                   \\ \hline
State12                                        & 14.508                                                                                             & \multicolumn{1}{l|}{\cellcolor[HTML]{EFEFEF}2.432 ± 0.562}  & 11                                                  & \multicolumn{1}{l|}{\cellcolor[HTML]{C0C0C0}2.423 ± 0.535} & \multicolumn{1}{l|}{\cellcolor[HTML]{C0C0C0}0.043 ± 0.006} & \multicolumn{1}{l|}{\cellcolor[HTML]{C0C0C0}124.028 ± 42.038} & 2                                                   & \multicolumn{1}{l|}{\cellcolor[HTML]{9B9B9B}/}               & \multicolumn{1}{l|}{\cellcolor[HTML]{9B9B9B}/}               & \multicolumn{1}{l|}{\cellcolor[HTML]{9B9B9B}/}                & 0                                                   \\ \hline
State13                                        & 13.170                                                                                             & \multicolumn{1}{l|}{\cellcolor[HTML]{EFEFEF}2.749 ± 0.755}  & 41                                                  & \multicolumn{1}{l|}{\cellcolor[HTML]{C0C0C0}2.463 ± 0.796} & \multicolumn{1}{l|}{\cellcolor[HTML]{C0C0C0}0.028 ± 0.010} & \multicolumn{1}{l|}{\cellcolor[HTML]{C0C0C0}156.544 ± 51.334} & 13                                                  & \multicolumn{1}{l|}{\cellcolor[HTML]{9B9B9B}/}               & \multicolumn{1}{l|}{\cellcolor[HTML]{9B9B9B}/}               & \multicolumn{1}{l|}{\cellcolor[HTML]{9B9B9B}/}                & 0                                                   \\ \hline
State14                                        & 23.357                                                                                             & \multicolumn{1}{l|}{\cellcolor[HTML]{EFEFEF}2.736 ± 0.401}  & 360                                                 & \multicolumn{1}{l|}{\cellcolor[HTML]{C0C0C0}2.701 ± 0.403} & \multicolumn{1}{l|}{\cellcolor[HTML]{C0C0C0}0.027 ± 0.013} & \multicolumn{1}{l|}{\cellcolor[HTML]{C0C0C0}140.076 ± 74.610} & 103                                                 & \multicolumn{1}{l|}{\cellcolor[HTML]{9B9B9B}/}               & \multicolumn{1}{l|}{\cellcolor[HTML]{9B9B9B}/}               & \multicolumn{1}{l|}{\cellcolor[HTML]{9B9B9B}/}                & 0                                                   \\ \hline
State15                                        & 22.209                                                                                             & \multicolumn{1}{l|}{\cellcolor[HTML]{EFEFEF}2.216 ± 0.342}  & 44                                                  & \multicolumn{1}{l|}{\cellcolor[HTML]{C0C0C0}2.216 ± 0.350} & \multicolumn{1}{l|}{\cellcolor[HTML]{C0C0C0}0.027 ± 0.014} & \multicolumn{1}{l|}{\cellcolor[HTML]{C0C0C0}182.852 ± 89.635} & 10                                                  & \multicolumn{1}{l|}{\cellcolor[HTML]{9B9B9B}/}               & \multicolumn{1}{l|}{\cellcolor[HTML]{9B9B9B}/}               & \multicolumn{1}{l|}{\cellcolor[HTML]{9B9B9B}/}                & 0                                                   \\ \hline
State16                                        & 19.387                                                                                             & \multicolumn{1}{l|}{\cellcolor[HTML]{EFEFEF}2.279 ± 0.177}  & 18                                                  & \multicolumn{1}{l|}{\cellcolor[HTML]{C0C0C0}2.376 ± 0.166} & \multicolumn{1}{l|}{\cellcolor[HTML]{C0C0C0}0.028 ± 0.005} & \multicolumn{1}{l|}{\cellcolor[HTML]{C0C0C0}165.186 ± 18.779} & 4                                                   & \multicolumn{1}{l|}{\cellcolor[HTML]{9B9B9B}/}               & \multicolumn{1}{l|}{\cellcolor[HTML]{9B9B9B}/}               & \multicolumn{1}{l|}{\cellcolor[HTML]{9B9B9B}/}                & 0                                                   \\ \hline
State17                                        & 16.003                                                                                             & \multicolumn{1}{l|}{\cellcolor[HTML]{EFEFEF}2.183 ± 0.262}  & 8                                                   & \multicolumn{1}{l|}{\cellcolor[HTML]{C0C0C0}2.271 ± 0.066} & \multicolumn{1}{l|}{\cellcolor[HTML]{C0C0C0}0.021 ± 0.015} & \multicolumn{1}{l|}{\cellcolor[HTML]{C0C0C0}176.574 ± 57.985} & 3                                                   & \multicolumn{1}{l|}{\cellcolor[HTML]{9B9B9B}/}               & \multicolumn{1}{l|}{\cellcolor[HTML]{9B9B9B}/}               & \multicolumn{1}{l|}{\cellcolor[HTML]{9B9B9B}/}                & 0                                                   \\ \hline
State18                                        & 37.067                                                                                             & \multicolumn{1}{l|}{\cellcolor[HTML]{EFEFEF}/}                & 0                                                   & \multicolumn{1}{l|}{\cellcolor[HTML]{C0C0C0}/}               & \multicolumn{1}{l|}{\cellcolor[HTML]{C0C0C0}/}               & \multicolumn{1}{l|}{\cellcolor[HTML]{C0C0C0}/}                  & 0                                                   & \multicolumn{1}{l|}{\cellcolor[HTML]{9B9B9B}/}               & \multicolumn{1}{l|}{\cellcolor[HTML]{9B9B9B}/}               & \multicolumn{1}{l|}{\cellcolor[HTML]{9B9B9B}/}                & 0                                                   \\ \hline
State19                                        & 41.184                                                                                             & \multicolumn{1}{l|}{\cellcolor[HTML]{EFEFEF}2.235 ± 0.000}  & 1                                                   & \multicolumn{1}{l|}{\cellcolor[HTML]{C0C0C0}2.235 ± 0.000} & \multicolumn{1}{l|}{\cellcolor[HTML]{C0C0C0}0.025 ± 0.000} & \multicolumn{1}{l|}{\cellcolor[HTML]{C0C0C0}355.061 ± 0.000}  & 1                                                   & \multicolumn{1}{l|}{\cellcolor[HTML]{9B9B9B}/}               & \multicolumn{1}{l|}{\cellcolor[HTML]{9B9B9B}/}               & \multicolumn{1}{l|}{\cellcolor[HTML]{9B9B9B}/}                & 0                                                   \\ \hline
\end{tabular}
    }
\begin{tablenotes}[flushleft]
    \small
        
          \item[a] a: Backbone energy of the reference state (in REU).
        
          \item[b] b: Based on RMSD only: \textit{Amount} indicates how many conformations are closest to each reference state; \textit{RMSD} reports their mean ± std.
        
          \item[c] c: Based on RMSD and dihedrals (\(U^2 < 0.2\), \(p > 0.05\)): same \textit{Amount} and \textit{RMSD} as [b];  \textit{EMD}, and \textit{Energy} reports their mean ± std.
        
          \item[d] d: Based on RMSD and dihedrals (\(U^2 < 0.2\), \(p > 0.05\)) and energy constraint (lower than the baseline).

  \end{tablenotes}

\end{sidewaystable}
\clearpage

\newpage
\section*{References}
{
\small
[1] Park, S., Khalili‑Araghi, F., Tajkhorshid, E.\ \& Schulten, K.\ (2003) Free energy calculation from steered molecular dynamics simulations using Jarzynski’s equality. {\it The Journal of Chemical Physics} {\bf 119}(6):3559–3566. American Institute of Physics.

[2] Park, S., Khalili‑Araghi, F.\ \& Strümpfer, J.\ (2015) {\it Stretching deca‑alanine}.

[3] Phillips, J.C., Hardy, D.J., Maia, J.D.C., Stone, J.E., Ribeiro, J.V., Bernardi, R.C., Buch, R., Fiorin, G., Hénin, J., Jiang, W.\ \emph{et al.}\ (2020) Scalable molecular dynamics on CPU and GPU architectures with NAMD. {\it The Journal of Chemical Physics} {\bf 153}(4). AIP Publishing.

[4] Eastman, P., Swails, J., Chodera, J.D., McGibbon, R.T., Zhao, Y., Beauchamp, K.A., Wang, L.‑P., Simmonett, A.C., Harrigan, M.P.\ \& Stern, C.D.\ (2017) OpenMM 7: Rapid development of high performance algorithms for molecular dynamics. {\it PLoS Computational Biology} {\bf 13}(7):e1005659. Public Library of Science San Francisco, CA USA.

[5] Chaudhury, S., Lyskov, S.\ \& Gray, J.J.\ (2010) PyRosetta: a script‑based interface for implementing molecular modeling algorithms using Rosetta. {\it Bioinformatics} {\bf 26}(5):689–691. Oxford University Press.

[6] Degiacomi, M.T.\ (2019) Coupling molecular dynamics and deep learning to mine protein conformational space. {\it Structure} {\bf 27}(6):1034–1040. Elsevier.
}
\end{document}